\documentclass[10pt,twocolumn,letterpaper]{article}

\usepackage[pagenumbers]{cvpr} % To force page numbers, e.g. for an arXiv version

\usepackage{graphicx}
\usepackage{amsmath}
\usepackage{amssymb}
\usepackage[english]{babel}
\usepackage{amsthm}
\usepackage{booktabs}
\usepackage{float}
\usepackage{stfloats}
\usepackage{color}
\usepackage{multirow}
\usepackage{amsmath,amsfonts,bm}

\newcommand{\ba}[1]{\begin{align}#1\end{align}}

\def\eqref#1{equation~\ref{#1}}
\def\1{\bm{1}}

\def\rvepsilon{{\boldsymbol{\epsilon}}}
\def\rvtheta{{\boldsymbol{\theta}}}
\def\rvphi{{\boldsymbol{\phi}}}
\def\rvmu{{\boldsymbol{\mu}}}
\def\rvpsi{{\boldsymbol{\psi}}}

\def\rvx{{\boldsymbol{x}}}
\def\rvy{{\boldsymbol{y}}}
\def\rvz{{\boldsymbol{z}}}

\def\vzero{{\bm{0}}}

\def\mI{{\bm{I}}}

\DeclareMathAlphabet{\mathsfit}{\encodingdefault}{\sfdefault}{m}{sl}
\SetMathAlphabet{\mathsfit}{bold}{\encodingdefault}{\sfdefault}{bx}{n}

\def\gN{{\mathcal{N}}}

\def\gT{{\mathcal{T}}}

\newcommand{\E}{\mathbb{E}}

\usepackage[pagebackref,breaklinks,colorlinks]{hyperref}

\usepackage[capitalize]{cleveref}
\crefname{section}{Sec.}{Secs.}
\Crefname{section}{Section}{Sections}
\Crefname{table}{Table}{Tables}
\crefname{table}{Tab.}{Tabs.}

\def\cvprPaperID{1579} % *** Enter the CVPR Paper ID here
\def\confName{CVPR}
\def\confYear{2023}

\begin{document}

\title{DR2: Diffusion-based Robust Degradation Remover for Blind Face Restoration}

\renewcommand{\thefootnote}{\fnsymbol{footnote}}

\author{%
Zhixin Wang\textsuperscript{1} \hspace{1cm} Xiaoyun Zhang\textsuperscript{1}\footnotemark[2]\hspace{1cm} Ziying Zhang\textsuperscript{1} \hspace{1cm}  Huangjie Zheng\textsuperscript{2} \\
Mingyuan Zhou\textsuperscript{2} \hspace{1cm} Ya Zhang\textsuperscript{1,3}  \hspace{1cm} Yanfeng Wang\textsuperscript{1,3}\footnotemark[2]\\
% Cooperative Medianet Innovation Center, 
\textsuperscript{1}Shanghai Jiao Tong University, \textsuperscript{3}Shanghai AI Laboratory, \textsuperscript{2}The University of Texas at Austin\\
{\tt\small \{dedsec\_z, xiaoyun.zhang, zyzhang2000, ya\_zhang, wangyanfeng\} @sjtu.edu.cn} \\
{\tt\small huangjie.zheng@utexas.edu, mingyuan.zhou@mccombs.utexas.edu} \\
}

\twocolumn[{%
\renewcommand\twocolumn[1][]{#1}%
\maketitle
\begin{center}
    \centering
    \includegraphics[width=1.00\linewidth, trim=0 20 0 25]{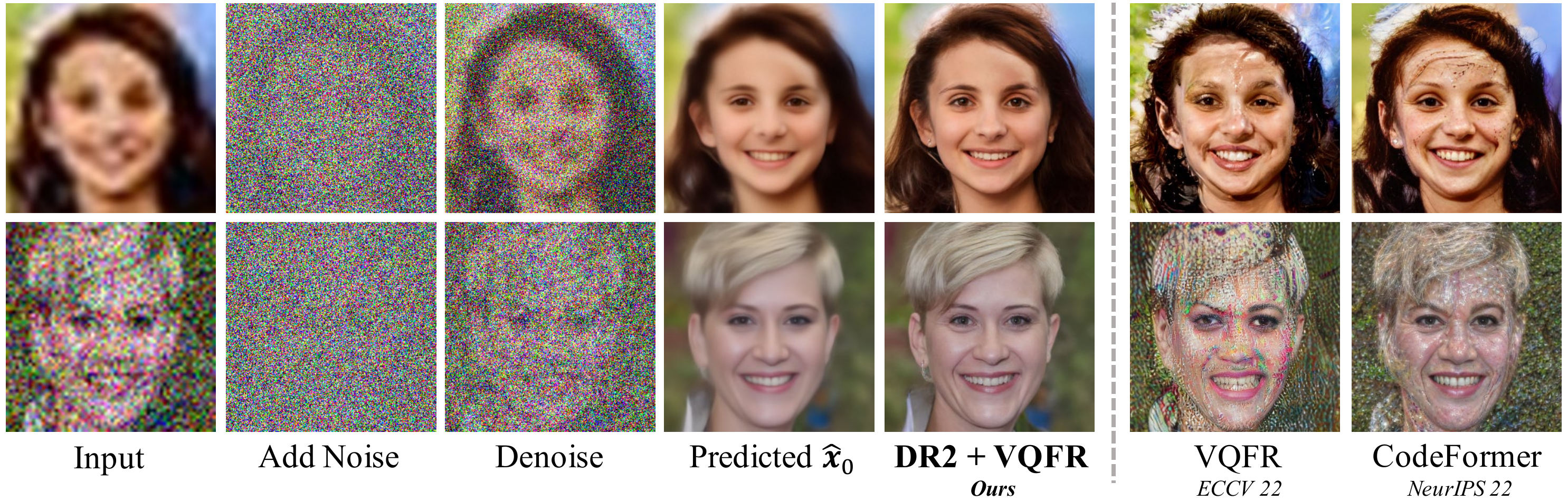}
    \captionof{figure}{\textbf{DR2 uses Denoising Diffusion Probabilistic Models to remove degradation}. The generative process is conditioned on the low-quality input after being diffused into a noisy status. As a result, DR2 predicts coarse faces $\hat{\rvx}_0$ regardless of the degradation type. On severely degraded images, our final restoration results achieve high quality with fewer artifacts than previous arts \cite{VQFR,CF}.}
    \label{Fig:1}
\end{center}%
}]
% \maketitle

%%%%%%%%% ABSTRACT
\begin{abstract}
\footnotetext[2]{Corresponding author.}
Blind face restoration usually synthesizes degraded low-quality data with a pre-defined degradation model for training, while more complex cases could happen in the real world. This gap between the assumed and actual degradation hurts the restoration performance where artifacts are often observed in the output. However, it is expensive and infeasible to include every type of degradation to cover real-world cases in the training data. To tackle this robustness issue, we propose \textbf{D}iffusion-based \textbf{R}obust \textbf{D}egradation \textbf{R}emover (DR2) to first transform the degraded image to a coarse but degradation-invariant prediction, then employ an enhancement module to restore the coarse prediction to a high-quality image. By leveraging a well-performing denoising diffusion probabilistic model, our DR2 diffuses input images to a noisy status where various types of degradation give way to Gaussian noise, and then captures semantic information through iterative denoising steps. As a result, DR2 is robust against common degradation (\eg blur, resize, noise and compression) and compatible with different designs of enhancement modules. Experiments in various settings show that our framework outperforms state-of-the-art methods on heavily degraded synthetic and real-world datasets.
\end{abstract}

%%%%%%%%% BODY TEXT
\section{Introduction}
\label{sec:intro}

Blind face restoration aims to restore high-quality face images from their low-quality counterparts suffering from unknown degradation, such as low-resolution \cite{GFPGAN13,GFPGAN48,GFPGAN09}, blur \cite{GFPGAN71}, noise\cite{GFPGAN39,GFPGAN58}, compression \cite{GFPGAN12}, \etc. Great improvement in restoration quality has been witnessed over the past few years with the exploitation of various facial priors. Geometric priors such as facial landmarks \cite{GFPGAN09}, parsing maps \cite{GFPGAN06,GFPGAN09}, and heatmaps \cite{GFPGAN69} are pivotal to recovering the shapes of facial components. Reference priors \cite{GFPGAN46,GFPGAN45,GFPGAN11} of high-quality images are used as guidance to improve details. Recent research investigates generative priors \cite{GFPGAN,GPEN} and high-quality dictionaries \cite{GFPGAN44,VQFR,CF}, which help to generate photo-realistic details and textures.

Despite the great progress in visual quality, these methods lack a robust mechanism to handle degraded inputs besides relying on pre-defined degradation to synthesize the training data. When applying them to images of severe or unseen degradation, undesired results with obvious artifacts can be observed. As shown in \cref{Fig:1}, artifacts typically appear when 1) the input image lacks high-frequency information due to downsampling or blur ($1^{st}$ row), in which case restoration networks can not generate adequate information, or 2) the input image bears corrupted high-frequency information due to noise or other degradation ($2^{nd}$ row), and restoration networks mistakenly use the corrupted information for restoration. The primary cause of this inadaptability is the inconsistency between the synthetic degradation of training data and the actual degradation in the real world.

% ---------------------------------------------------------------------------------------------------------------

\begin{figure}
    \centering
    \includegraphics[width=1.0\columnwidth, trim=0 15 0 0]{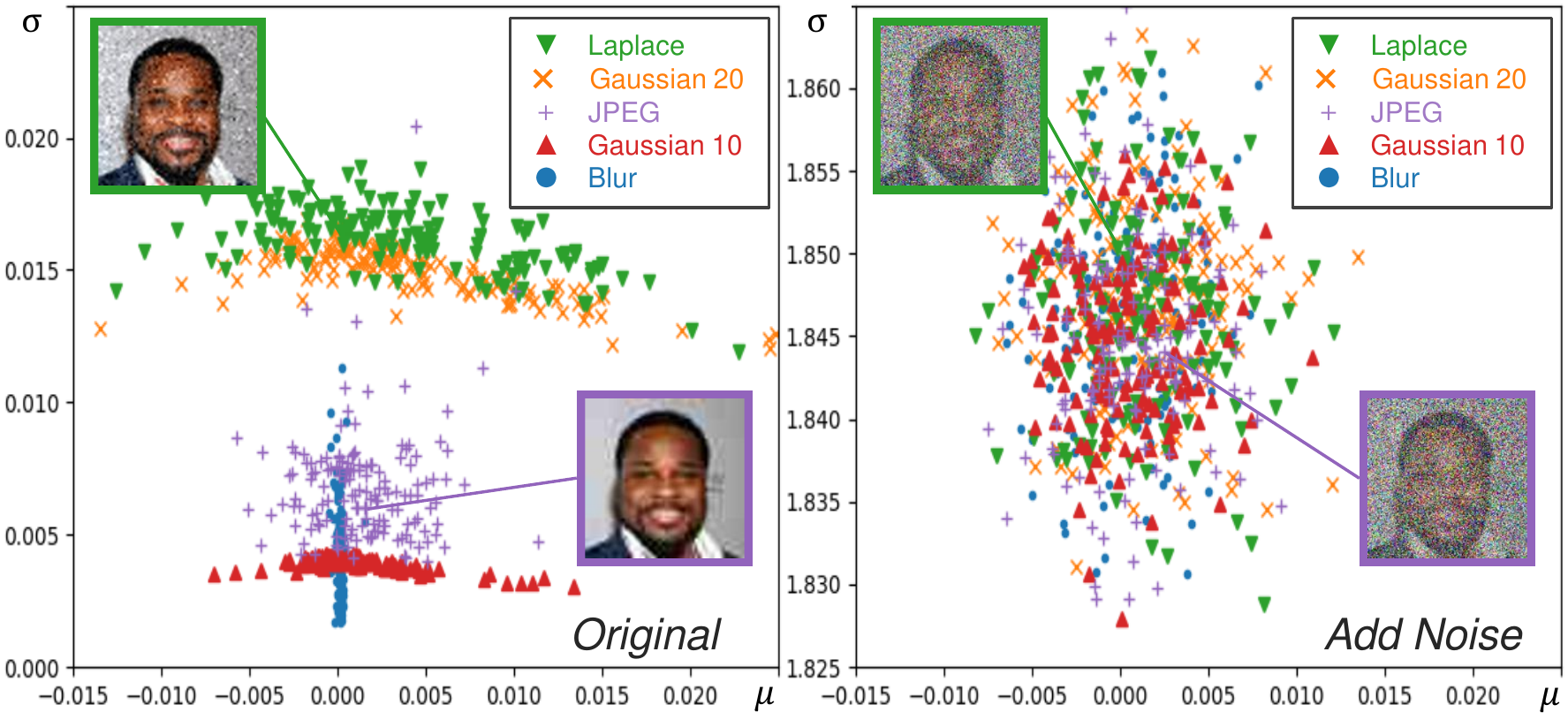}
    \caption{\textbf{Mean and standard variation of pixel-wise error distribution.} \textbf{\textit{(Left)}} the error between original degraded input $\rvy$ and its ground truth low-resolution image $\hat{\rvy}$ (only bicubically downsampled); \textbf{\textit{(Right)}}  the error between $q(\rvy_{500} | \rvy)$ and $q(\hat{\rvy}_{500} | \hat{\rvy})$ sampled by \cref{Eq:forw_detail}, with extra Gaussian noise added by the diffusion function.}
    \label{Fig:2}
\end{figure}

Expanding the synthetic degradation model for training would improve the models’ adaptability but it is apparently difficult and expensive to simulate every possible degradation in the real world. To alleviate the dependency on synthetic degradation, we leverage a well-performing denoising diffusion probabilistic model (DDPM) \cite{ILVR14,ILVR39} to remove the degradation from inputs. DDPM generates images through a stochastic iterative denoising process and Gaussian noisy images can provide guidance to the generative process \cite{ILVR,Repaint}. As shown in \cref{Fig:2}, noisy images are \textbf{degradation-irrelevant} conditions for DDPM generative process. Adding extra Gaussian noise (right) makes different degradation less distinguishable compared with the original distribution (left), while DDPM can still capture the semantic information within this noise status and recover clean face images. This property of pretrained DDPM makes it a robust degradation removal module though only high-quality face images are used for training the DDPM.

Our overall blind face restoration framework DR2E consists of the 
\textbf{D}iffusion-based \textbf{R}obust \textbf{D}egradation \textbf{R}emover (DR2) and an \textbf{E}nhancement module. In the first stage, DR2 first transforms the degraded images into coarse, smooth, and visually clean intermediate results, which fall into a degradation-invariant distribution ($4^{th}$ column in \cref{Fig:1}). In the second stage, the degradation-invariant images are further processed by the enhancement module for high-quality details. By this design, the enhancement module is compatible with various designs of restoration methods in seeking the best restoration quality, ensuring our DR2E achieves both strong robustness and high quality.

% ---------------------------------------------------------------------------------------------------------------

We summarize the contributions as follows. 
% (1) We propose DR2E, a two-stage blind face restoration framework that consists of a Diffusion-based Robust Degradation Remover (DR2) and an enhancement module, achieving both strong robustness and high quality. 
(1) We propose DR2 that leverages a pretrained diffusion model to remove degradation, achieving robustness against complex degradation without using synthetic degradation for training. 
(2) Together with an enhancement module, we employ DR2 in a two-stage blind face restoration framework, namely DR2E. The enhancement module has great flexibility in incorporating a variety of restoration methods to achieve high restoration quality.
% We systematically show the advantage of decomposing the face restoration framework in two stages by proposing DR2E that consists of DR2 and an enhancement module. Such decomposition ensures DR2 can leverage a pre-trained diffusion model to remove degradation information while preserving semantic information. Plus the enhancement module, the degradation-invariant image can be further completed as a high-quality image. Both modules have great flexibility to incorporating state-of-the-art diffusion models and restoration methods.
(3) Comprehensive studies and experiments show that our framework outperforms state-of-the-art methods on heavily degraded synthetic and real-world datasets.

% ---------------------------------------------------------------------------------------------------------------

\section{Related Work}
\label{sec:related}

\begin{figure*}
\centering
\includegraphics[width=1.0\linewidth, trim=0 10 0 0]{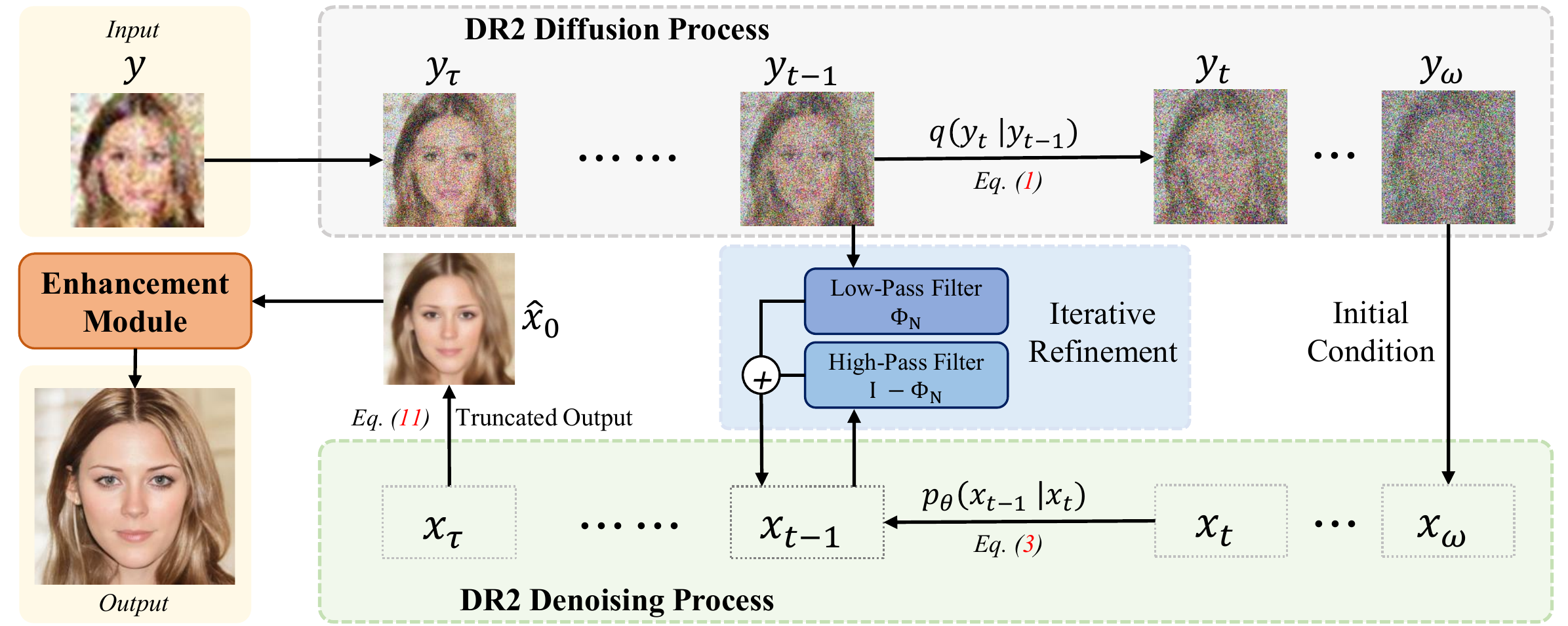}
\caption{\textbf{Overall DR2E framework}. It consists of DR2 as the degradation removal module and an enhancement module. During inference, we sample $\rvy_\tau$, $\rvy_{\tau+1}$,...,$\rvy_{\omega}$ through diffusion process and use them as guidance. We use $\rvy_{\omega}$ as $\rvx_{\omega}$ and start the denoising process from step $\omega$ to $\tau$. After each transition from $\rvx_t$ to $\rvx_{t-1}$, we combine the low-frequency of $\rvy_{t-1}$ and high-frequency of $\rvx_{t-1}$. At step $\tau$, we predict $\hat{\rvx}_0$ based on the estimated noise. Then the enhancement module produces high-quality output from $\hat{\rvx}_0$.}
\label{Fig:3}
\end{figure*}

% ---------------------------------------------------------------------------------------
% ---------------------------------- Blind Face Restoration -----------------------------
% ---------------------------------------------------------------------------------------

\noindent \textbf{Blind Face Restoration} Based on face hallucination or face super-resolution \cite{GFPGAN05,GFPGAN30,GFPGAN66,GFPGAN70}, blind face restoration aims to restore high-quality faces from low-quality images with unknown and complex degradation. Many facial priors are exploited to alleviate dependency on degraded inputs. Geometry priors, including facial landmarks \cite{GFPGAN09,GFPGAN37,GFPGAN77}, parsing maps \cite{GFPGAN58,GFPGAN06,GFPGAN09}, and facial component heatmaps \cite{GFPGAN69} help to recover accurate shapes but contain no information on details in themselves. Reference priors \cite{GFPGAN46,GFPGAN45,GFPGAN11} of high-quality images are used to recover details or preserve identity. To further boost restoration quality, generative priors like pretrained StyleGAN \cite{GFPGAN35,GFPGAN36} are used to provide vivid textures and details. PULSE \cite{GFPGAN52} uses latent optimization to find latent code of high-quality face, while more efficiently, GPEN \cite{GPEN}, GFP-GAN \cite{GFPGAN}, and GLEAN \cite{CF02} embed generative priors into the encoder-decoder structure. Another category of methods utilizes pretrained Vector-Quantize \cite{VQFR30,VQFR11,VQFR27} codebooks. DFDNet \cite{GFPGAN44} suggests constructing dictionaries of each component (\eg eyes, mouth), while recent VQFR \cite{VQFR} and CodeFormer \cite{CF} pretrain high-quality dictionaries on entire faces, acquiring rich expressiveness.

% Despite these priors, most aforementioned methods rely on synthetically degraded datasets for training. This leads to performance drops if the testing degradation setting is not aligned with training.

% ---------------------------------------------------------------------------------------
% ------------------------- Denoising Diffusion Probabilistic Models --------------------
% ---------------------------------------------------------------------------------------

\noindent \textbf{Diffusion Models} Denoising Diffusion Probabilistic Models (DDPM) \cite{ILVR14,ILVR39} are a fast-developing class of generative models in unconditional image generation rivaling Generative Adversarial Networks (GAN) \cite{GAN,SR3_19,SR3_36}. Recent research utilizes it for super-resolution. SR3 \cite{SR3_} modifies DDPM to be conditioned on low-resolution images through channel-wise concatenation. However, it fixes the degradation to simple downsampling and does not apply to other degradation settings. Latent Diffusion \cite{LDM} performs super-resolution in a similar concatenation manner but in a low-dimensional latent space. ILVR \cite{ILVR} proposes a conditioning method to control the generative process of pretrained DDPM for image-translation tasks. Diffusion-based methods face a common problem of slow sampling speed, while our DR2E adopts a hybrid architecture like \cite{TDPM} to speed up the sampling process.

\section{Methodology}
\label{sec:method}

Our proposed DR2E framework is depicted in \cref{Fig:3}, which consists of the degradation remover DR2 and an enhancement module. Given an input image $\rvy$ suffering from unknown degradation, diffused low-quality information $\rvy_{t-1}$ is provided to refine the generative process. As a result, DR2 recovers a coarse result $\hat{\rvx}_0$ that is semantically close to $\rvy$ and degradation-invariant. Then the enhancement module maps $\hat{\rvx}_0$ to the final output with higher resolution and high-quality details.

% ================================================================================================
% ============ 3.1 Preliminaries: Denoising Diffusion Probabilistic Models =======================
% ================================================================================================
\subsection{Preliminary}%: Denoising Diffusion Probabilistic Models}
\label{Sec:3.1}

Denoising Diffusion Probabilistic Models (DDPM) \cite{ILVR14,ILVR39} are a class of generative models that first pre-defines a variance schedule $\{\beta_1, \beta_2,...,\beta_T\}$ to progressively corrupt an image $\rvx_0$ to a noisy status through forward (diffusion) process:

\begin{equation}
    q(\rvx_t | \rvx_{t-1}) = \gN(\rvx_t; \sqrt{1 - \beta_t}\rvx_{t-1}, \beta_t \mI)
    \label{Eq:forward}
\end{equation}

Moreover, based on the property of the Markov chain, for any intermediate timestep $t\in\{1,2,...,T\}$, the corresponding noisy distribution has an analytic form:
\begin{align}
    q(\rvx_t|\rvx_0) &= \gN(\rvx_t; \sqrt{\bar \alpha_t}\rvx_{0},   (1 - \bar \alpha_t)\mI) \notag \\
                     &= \sqrt{\bar \alpha_t} \rvx_0 + \sqrt{1 - \bar \alpha_t} \rvepsilon
    \label{Eq:forw_detail}
\end{align}

\noindent where $\bar{\alpha}_t := \prod_{s=1}^t (1 -\beta_s)$ and $\rvepsilon \sim \gN(\vzero, \mI)$. Then $\rvx_T \sim \gN(\vzero, \mI)$ if $T$ is big enough, usually $T = 1000$.

The model progressively generates images by reversing the forward process. The generative process is also a Gaussian transition with the learned mean $\rvmu_\theta$:

\begin{equation}
    p_\theta(\rvx_{t-1} | \rvx_t) = \gN(\rvx_{t-1}; \rvmu_\theta(\rvx_t, t), \sigma_t^2 \mI)
    \label{Eq:reverse}
\end{equation}

\noindent where $\sigma_t$ is usually a pre-defined constant related to the variance schedule, and $\rvmu_\theta(\rvx_t, t)$ is usually parameterized by a denoising U-Net $\rvepsilon_\theta(\rvx_t, t)$ \cite{GFPGAN56} with the following equivalence:
\begin{equation}
    \rvmu_\theta(\rvx_t, t) = \frac{1}{\sqrt{\alpha_t}}(\rvx_t - \frac{1 - \alpha_t}{\sqrt{1 - \bar{\alpha}_t}} \rvepsilon_\theta(\rvx_t, t))
    \label{Eq:rev_detail}
\end{equation}

% ================================================================================================
% ===================== 3.2 Framework overview ===========================
% ================================================================================================
\subsection{Framework Overview}
\label{Sec:3.2}

Suppose the low-quality image $\rvy$ is degraded from the high-quality ground truth $\rvx \sim \mathcal{X}(\rvx)$ as $\rvy = \gT(\rvx, \rvz)$ where $\rvz$ describes the degradation model. Previous studies constructs the inverse function $\gT^{-1}(\cdot, \rvz)$ by modeling $p(\rvx| \rvy, \rvz)$ with a pre-defined $\rvz$ \cite{GFPGAN13,GFPGAN48,GFPGAN09}. It meets the adaptation problem when actual degradation $\rvz^\prime$ in the real world is far from $\rvz$.

To overcome this challenge, we propose to model $p(\rvx| \rvy)$ without a known $\rvz$ by a two-stage framework: it first removes degradation from inputs and get $\hat{x}_0$, then maps degradation-invariant $\hat{x}_0$ to high-quality outputs. Our target is to maximize the likelihood:
\ba{
\quad p_{\rvpsi,\rvphi}(\rvx| \rvy) &= \int p_\rvpsi(\rvx|\hat{\rvx}_0) p_\rvphi(\hat{\rvx}_0|\rvy) d\hat{\rvx}_0 \notag \\
&= \E_{\hat{\rvx}_0 \sim p_\rvphi(\hat{\rvx}_0|\rvy)} \left[ p_\rvpsi(\rvx|\hat{\rvx}_0)\right] 
\label{eq:objective},
} 

$p_\rvphi(\hat{\rvx}_0|\rvy)$ corresponds to the degradation removal module, and $p_\rvpsi(\rvx|\hat{\rvx}_0)$ corresponds to the enhancement module. For the first stage, instead of directly learning the mapping from $\rvy$ to $\hat{\rvx}_0$ which usually involves a pre-defined degradation model $\rvz$, we come up with an important assumption and propose a diffusion-based method to remove degradation.

\noindent \textbf{Assumption.} For the diffusion process defined in \cref{Eq:forw_detail}, (1) there exists an intermediate timestep $\tau$ such that for $t > \tau$, the distance between $q(\rvx_t|\rvx)$ and $q(\rvy_t|\rvy)$ is close especially in the low-frequency part; (2) there exists $\omega > \tau$ such that the distance between $q(\rvx_\omega|\rvx)$ and $q(\rvy_\omega|\rvy)$ is eventually small enough, satisfying ${q(\rvx_\omega|\rvx)} \approx {q(\rvy_\omega|\rvy)}$.

Note this assumption is not strong, as paired $\rvx$ and $\rvy$ would share similar low-frequency contents, and for sufficiently large $t \approx T$, $q(\rvx_t|\rvx)$ and $q(\rvy_t|\rvy)$ are naturally close to the standard $\gN(\vzero, \mI)$. This assumption is also qualitatively justified in \cref{Fig:2}. Intuitively, if $\rvx$ and $\rvy$ are close in distribution (implying mild degradation), we can find $\omega$ and $\tau$ in a relatively small value and vice versa. 

Then we rewrite the objective of the degradation removal module by applying the assumption ${q(\rvx_\omega|\rvx)} \approx {q(\rvy_\omega|\rvy)}$:
\ba{
\label{Eq:6} p_\rvphi(\hat{\rvx}_0|\rvy) &= \int p(\hat{\rvx}_0|\rvx_\tau) p_\theta(\rvx_\tau|\rvy_\omega) q(\rvy_\omega | \rvy) d\rvx_\tau d\rvy_\omega \\
\label{Eq:7} & \approx \int p(\hat{\rvx}_0|\rvx_\tau) p_\theta(\rvx_\tau|\rvx_\omega) q(\rvx_\omega | \rvx) d\rvx_\tau d\rvx_\omega  \\
\label{Eq:8}  & p_\theta(\rvx_\tau|\rvx_\omega) = \prod_{t={\tau+1}}^{\omega} p_\theta(\rvx_{t-1}|\rvx_t)
}

By replacing variable from $\rvy_\omega$ to $\rvx_\omega$, \cref{Eq:7} and \cref{Eq:8} naturally yields a DDPM model that denoises $\rvx_\omega$ back to $\rvx_\tau$, and we can further predict $\hat{\rvx}_0$ by the reverse of \cref{Eq:forw_detail}. $\hat{\rvx}_0$ would maintain semantics with $\rvx$ if proper conditioning methods like \cite{ILVR} is adopted. So by leveraging a DDPM, we propose Diffusion-based Robust Degradation Remover (DR2) according to  \cref{Eq:6}.

% ================================================================================================
% ===================== 3.3 Diffusion-based Robust Degradation Remover ===========================
% ================================================================================================
\subsection{Diffusion-based Robust Degradation Remover}
\label{Sec3:3}

Consider a pretrained DDPM $p_\theta(\rvx_{t-1}|\rvx_t)$ (\cref{Eq:reverse}) with a denoising U-Net $\rvepsilon_\theta(\rvx_t, t)$ pretrained on high-quality face dataset. We respectively implement $q(\rvy_\omega | \rvy)$, $p_\theta(\rvx_\tau|\rvy_\omega)$ and $p(\hat{\rvx}_0|\rvx_\tau)$ in \cref{Eq:6} by three steps in below.

% ----------------------------------------------------------------------
% -------------------------- Initial Condition -------------------------
% ----------------------------------------------------------------------

\noindent \textbf{(1) Initial Condition at $\mathbf{\omega}$.} We first ``forward" the degraded image $\rvy$ to an initial condition $\rvy_\omega$ by sampling from \cref{Eq:forw_detail}  and use it as $\rvx_\omega$: 

\begin{equation}
     \rvx_\omega := \rvy_{\omega} = \sqrt{\bar{\alpha}_{\omega}} \rvy + \sqrt{1 - \bar{\alpha}_{\omega}} \rvepsilon,
     \label{Eq:init_cond}
\end{equation}

\noindent $\omega \in \{1,2,...,T\}$. This corresponds to $q(\rvx_\omega|\rvy)$ in \cref{Eq:6}. Then the DR2 denoising process starts at step $\omega$. This reduces the samplings steps and helps to speed up as well.

% ----------------------------------------------------------------------
% ------------------------ Iterative Refinement ------------------------
% ----------------------------------------------------------------------
\noindent \textbf{(2) Iterative Refinement.} After each transition from $\rvx_t$ to $\rvx_{t-1}$ ($\tau + 1 \leqslant t \leqslant \omega$), we sample $\rvy_{t-1}$ from $\rvy$ through \cref{Eq:forw_detail}. Based on Assumption (1), we replace the low-frequency part of $\rvx_{t-1}$ with that of $\rvy_{t-1}$ because they are close in distribution, which is fomulated as:

\begin{equation}
        \rvx_{t-1} := \Phi_N(\rvy_{t-1}) + (\mathbf{I} - \Phi_N)(\rvx_{t-1})
        \label{Eq:ILVR}
\end{equation}

\noindent where $\Phi_N(\cdot)$ denotes a low-pass filter implemented by downsampling and upsampling the image with a sharing scale factor $N$. We drop the high-frequency part of $\rvy$ for it contains little information due to degradation. Unfiltered degradation that remained in the low-frequency part would be covered by the added noise. These conditional denoising steps correspond to $p_\rvtheta(\rvx_\tau|\rvy_\omega)$ in \cref{Eq:6}, which ensure the result shares basic semantics with $y$. 

Iterative refinement is pivotal for preserving the low-frequency information of the input images. With the iterative refinement, the choice of $\omega$ and the randomness of Gaussian noise affect little to the result. We present ablation study in the supplementary for illustration.

% ----------------------------------------------------------------------
% ---------------------- Truncated Output ---------------------
% ----------------------------------------------------------------------

\noindent \textbf{(3) Truncated Output at $\mathbf{\tau}$.} As $t$ gets smaller, the noise level gets milder and the distance between $q(\rvx_t|\rvx)$ and $q(\rvy_t|\rvy)$ gets larger. For small $t$, the original degradation is more dominating in $q(\rvy_t|\rvy)$ than the added Gaussian noise. So the denoising process is truncated before $t$ is too small. We use predicted noise at step $\tau$ $(0 < \tau < \omega)$ to estimate the generation result as follows:

\begin{equation}
    \hat{\rvx}_0 = \frac{1}{\sqrt{\bar \alpha_\tau}} (\rvx_\tau - \sqrt{1 - \bar \alpha_\tau} \rvepsilon_\theta(\rvx_\tau, \tau))
\end{equation}
This corresponds to $p(\hat{\rvx}_0|\rvx_\tau)$ in \cref{Eq:6}. $\hat{\rvx}_0$ is the output of DR2, which maintains the basic semantics of $\rvy$ and is removed from various degradation. 

% To sum up, the robustness of DR2 attributes to 1) only using low-frequency information (controlled by $N$) of the input for conditioning and 2) adding variant levels of Gaussian noise (controlled by $\tau$) on the input to cover the original degradation. 

\noindent \textbf{Selection of $N$ and $\tau$.} Downsampling factor $N$ and output step $\tau$ have significant effects on the fidelity and ``cleanness" of $\hat{\rvx}_0$. We conduct ablation studies in \cref{Sec:4.4} to show the effects of these two hyper-parameters. The best choices of $N$ and $\tau$ are data-dependent. Generally speaking, big $N$ and $\tau$ are more effective to remove the degradation but lead to lower fidelity. On the contrary, small $N$ and $\tau$ leads to high fidelity, but may keep the degradation in the outputs. While $\omega$ is empirically fixed to $\tau + 0.25T$. 

% ================================================================================================
% =================================== 3.4 Enhancement Module ======================================
% ================================================================================================

\subsection{Enhancement Module}
\label{Sec:3.4}

With outputs of DR2, restoring the high-quality details only requires training an enhancement module $p_\rvpsi(\rvx|\hat{\rvx}_0)$ (\cref{eq:objective}). Here we do not hypothesize about the specific method or architecture of this module. Any neural network that can be trained to map a low-quality image to its high-quality counterpart can be plugged in our framework. And the enhancement module is independently trained with its proposed loss functions.

\noindent \textbf{Backbones.} In practice, without loss of generality, we choose SPARNetHD \cite{SPAR} that utilized no facial priors, and VQFR \cite{VQFR} that pretrain a high-quality VQ codebook \cite{VQFR30,VQFR11,VQFR27} as two alternative backbones for our enhancement module to justify that it can be compatible with a broad choice of existing methods. We denote them as DR2 + SPAR and DR2 + VQFR respectively.

\noindent \textbf{Training Data.} Any pretrained blind face restoration models can be directly plugged-in without further finetuning, but in order to help the enhancement module adapt better and faster to DR2 outputs, we suggest constructing training data for the enhancement module using DR2 as follows: 

\begin{equation}
    % \vspace{-1mm}
    \rvy = DR2(\rvx;N,\tau) \circledast k_\sigma
    \label{Eq:cond_aug}
\end{equation}

Given a high-quality image $\rvx$, we first use DR2 to reconstruct itself with controlling parameters $(N, \tau)$ then convolve it with an Gaussian blur kernel $k_\sigma$. This helps the enhancement module adapt better and faster to DR2 outputs, which is recommended but not compulsory. Noting that beside this augmentation, \textbf{no} other degradation model is required in the training process as what previous works \cite{SPAR,GPEN,GFPGAN,CF,VQFR} do by using \cref{Eq:deg}.

\section{Experiments}
\label{sec:exp}

\begin{figure*}
\centering
\includegraphics[width=1.0\linewidth, trim=0 15 0 10 ]{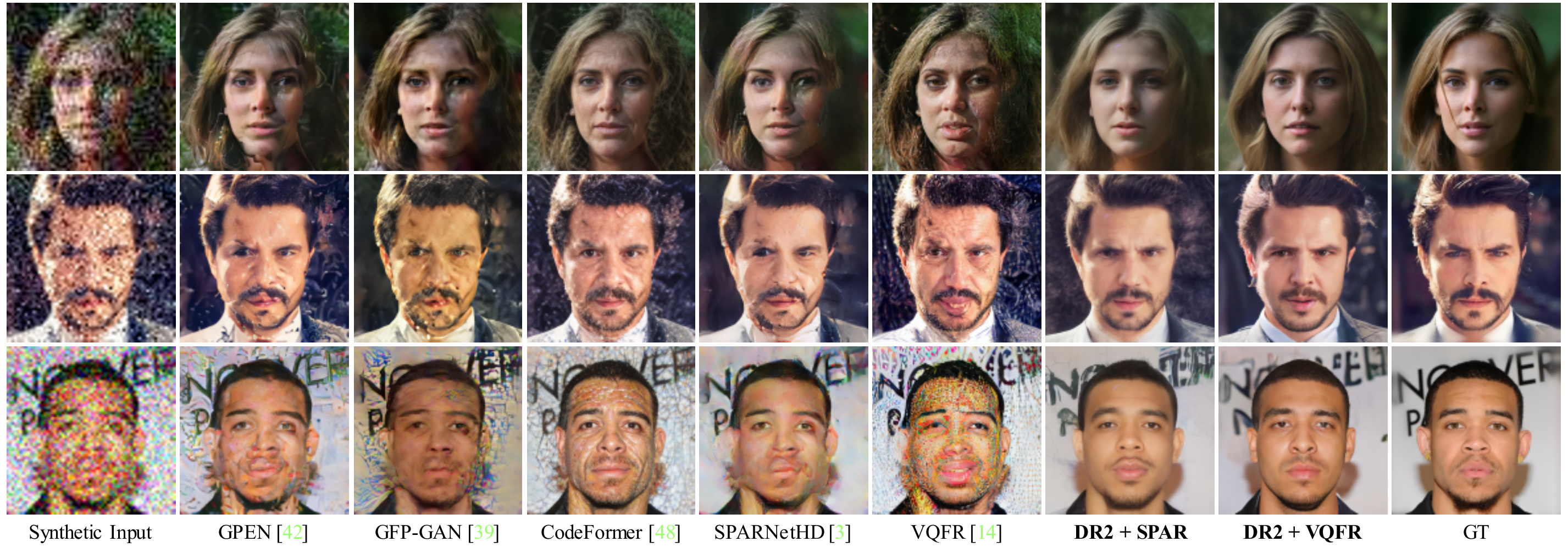}
\caption{\textbf{Qualitative comparison on the CelebA-Test dataset.} Our method with different enhancement module backbones achieve higher restoration quality with fewer artifacts despite the heavy degradation in inputs.}
\label{Fig:4}
\end{figure*}

%============================================================================================
%=============================== 4.1 Datasets and Implementation ============================
%============================================================================================
\subsection{Datasets and Implementation}
\label{sec:4.1}
\noindent \textbf{Implementation.} DR2 and the enhancement module are independently trained on FFHQ dataset \cite{GFPGAN35}, which contains 70,000 high-quality face images. We use pretrained DDPM proposed by \cite{ILVR} for our DR2. As introduced in \cref{Sec:3.4}, we choose SPARNetHD \cite{SPAR} and VQFR \cite{VQFR} as two alternative architectures for the enhancement module. We train SPARNetHD backbone from scratch with training data constructed by \cref{Eq:cond_aug}. We set $N = 4$ and randomly sample $\tau$, $\sigma$ from $\{50, 100, 150, 200\}$, $\{1 : 7\}$, respectively. As for VQFR backbone, we use its official pretrained model. 

\noindent \textbf{Testing Datasets}. We construct one synthetic dataset and four real-world datasets for testing. A brief introduction of each is as followed:

$\bullet$ \textit{CelebA-Test}. Following previous works \cite{SPAR,GPEN,GFPGAN,CF,VQFR}, we adopt a commonly used degradation model as follows to synthesize testing data from CelebA-HQ \cite{GFPGAN51}:
\begin{equation}
    \vspace{-0.5mm}
    \rvy = [(\rvx \circledast k_\sigma) \downarrow_r + n_\delta]_{JPEG_q}
    \label{Eq:deg}
\end{equation}

\noindent A high-quality image $\rvx$ is first convolved with a Gaussian blur kernel $k_\sigma$, then bicubically downsampled with a scale factor $r$. $n_\delta$ represents additive noise and is randomly chosen from Gaussian, Laplace, and Poisson. Finally, JPEG compression with quality $q$ is applied. We use $r$ = 16, 8, and 4 to form three restoration tasks denoted as $\mathbf{16\times}$, $\mathbf{8\times}$, and $\mathbf{4\times}$. For each upsampling factor, we generate three splits with different levels of degradation and each split contains 1,000 images. The \textit{mild} split randomly samples $\sigma$, $\delta$ and $q$ from $\{3:5\}$, $\{5:20\}$, $\{60:80\}$, respectively. The \textit{medium} from $\{5:7\}$, $\{15:40\}$, $\{40:60\}$. And the \textit{severe} split from $\{7:9\}$, $\{25:50\}$, $\{30:40\}$.

$\bullet$ \textit{WIDER-Normal} and \textit{WIDER-Critical}. We select 400 critical cases suffering from heavy degradation (mainly low-resolution) from WIDER-face dataset \cite{CF41} to form the WIDER-Critical dataset and another 400 regular cases for WIDER-Normal dataset.

$\bullet$ \textit{CelebChild} contains 180 child faces of celebrities collected from the Internet. Most of them are only mildly degraded.

$\bullet$ \textit{LFW-Test}. LFW \cite{GFPGAN29} contains low-quality images with mild degradation from the Internet. We choose 1,000 testing images of different identities.

During testing, we conduct grid search for best controlling parameters $(N,\tau)$ of DR2 for each dataset. Detailed parameter settings are presented in the suplementary.

%============================================================================================
%====================== 4.2 Comparisons with State-of-the-art Methods =======================
%============================================================================================
\subsection{Comparisons with State-of-the-art Methods}
\label{sec:4.2}

We compare our method with several state-of-the-art face restoration methods: DFDNet \cite{GFPGAN44}, SPARNetHD \cite{SPAR}, GFP-GAN \cite{GFPGAN}, GPEN \cite{GPEN}, VQFR \cite{VQFR}, and Codeformer \cite{CF}. We adopt their \textit{official} codes and pretrained models.

For evaluation, we adopt pixel-wise metrics (PSNR and SSIM) and the perceptual metric (LPIPS \cite{GFPGAN73}) for the CelebA-Test with ground truth. We also employ the widely-used non-reference perceptual metric FID \cite{GFPGAN27}.

\noindent \textbf{Synthetic CelebA-Test.} For each upsampling factor, we calculate evaluation metrics on three splits and present the average in \cref{Tab:1}. For $16\times$ and $8\times$ upsampling tasks where degradation is severe due to low resolution, DR2 + VQFR and DR2 + SPAR achieve the best and the second-best LPIPS and FID scores, indicating our results are perceptually close to the ground truth. Noting that DR2 + VQFR is better at perceptual metrics (LPIPS and FID) thanks to the pretrained high-quality codebook, and DR2 + SPAR is better at pixel-wise metrics (PSNR and SSIM) because without facial priors, the outputs have higher fidelity to the inputs. For $4\times$ upsampling task where degradation is relatively milder, previous methods trained on similar synthetic degradation manage to produce high-quality images without obvious artifacts. But our methods still obtain superior FID scores, showing our outputs have closer distribution to ground truth on different settings.

\begin{table*}
  \centering
  \resizebox{\textwidth}{24mm}{
      \begin{tabular}{c|c c c c|c c c c|c c c c}
        \toprule
         & \multicolumn{4}{c|}{$\times16$} & \multicolumn{4}{c|}{$\times8$} & \multicolumn{4}{c}{$\times4$} \\
         Methods & LPIPS$\downarrow$ & FID$\downarrow$ & PSNR$\uparrow$ & SSIM$\uparrow$ & LPIPS$\downarrow$ & FID$\downarrow$ & PSNR$\uparrow$ & SSIM$\uparrow$ & LPIPS$\downarrow$ & FID$\downarrow$ & PSNR$\uparrow$ & SSIM$\uparrow$ \\
        \midrule
         DFDNet* \cite{GFPGAN44}     & 0.5511 & 109.41 & 20.80 & 0.4929 & 0.5033 & 120.13 & 21.75 & 0.4758 & 0.4405 & 98.10 & 23.81 & 0.5357 \\
         GPEN \cite{GPEN}     & 0.4313 &  81.57 & 21.77 & 0.5916 & 0.3745 & 64.00  & 24.02 & 0.6398 & 0.2934 & 53.56 & \textcolor{blue}{\textit{26.38}} & 0.7057 \\
         GFP-GAN \cite{GFPGAN}    & 0.5430 & 139.13 & 18.35 & 0.4578 & 0.3233 & 56.88  & 23.36 & 0.6695 & 0.2720 & 58.78 & 24.94 & 0.7244 \\
         CodeFormer \cite{CF}   & 0.5176 & 117.17 & 19.70 & 0.4553 & 0.3465 & 71.22  & 23.04 & 0.5950 & \textcolor{red}{\textbf{0.2587}} & 61.41 & 26.33 & 0.7065 \\
        \midrule
         SPARNetHD \cite{SPAR}    & 0.4289 & 77.02  & \textcolor{blue}{\textit{22.28}} & 0.6114 & 0.3361 & 59.66 & \textcolor{blue}{\textit{24.71}} & 0.6743 & 0.2638 & 53.20 & \textcolor{red}{\textbf{26.59}} & \textcolor{blue}{\textit{0.7255}} \\
         VQFR \cite{VQFR}        & 0.6312 & 152.56 & 17.73 & 0.3381 & 0.4214 & 66.54 & 21.83 & 0.5345 & 0.3094 & 52.39 & 23.52 & 0.6335 \\
        \midrule
         \textbf{DR2 + SPAR(ours)}  & \textcolor{blue}{\textit{0.3908}} & \textcolor{blue}{\textit{53.22}} & \textcolor{red}{\textbf{22.29}} & \textcolor{red}{\textbf{0.6587}} & \textcolor{blue}{\textit{0.3218}} & \textcolor{blue}{\textit{56.29}} & \textcolor{red}{\textbf{24.78}} & \textcolor{red}{\textbf{0.6966}} & \textcolor{blue}{\textit{0.2635}} & \textcolor{blue}{\textit{51.44}} & 26.28 & \textcolor{red}{\textbf{0.7263}} \\
         \textbf{DR2 + VQFR(ours)}  & \textcolor{red}{\textbf{0.3893}} & \textcolor{red}{\textbf{47.29}} & 21.29 & \textcolor{blue}{\textit{0.6222}} & \textcolor{red}{\textbf{0.3167}} & \textcolor{red}{\textbf{53.82}} & 23.40 & \textcolor{blue}{\textit{0.6802}} & 0.2902 & \textcolor{red}{\textbf{51.41}} & 24.04 & 0.6844 \\
        \bottomrule
      \end{tabular}
  }
  \caption{Quantitative comparisons on \textbf{CelebA-Test} dataset. \textcolor{red}{\textbf{Red}} and \textcolor{blue}{\textit{blue}} indicates the best and the second best performance. '*' denotes using ground-truth facial landmarks as input.}
  \label{Tab:1}
\end{table*}

Qualitative comparisons from are presented in \cref{Fig:4}. Our methods produce fewer artifacts on severely degraded inputs compared with previous methods.

\begin{table}
  \centering
  \resizebox{\columnwidth}{23mm}{
      \begin{tabular}{c|c c c c}
        \toprule
        Datasets & W-Cr & W-Nm & Celeb-C & LFW \\
         Methods & FID$\downarrow$ & FID$\downarrow$ & FID$\downarrow$ & FID$\downarrow$ \\
        \midrule
         DFDNet \cite{GFPGAN44}           & 78.87 & 73.12 & 107.18 & 64.89 \\
         GPEN \cite{GPEN}             & 65.06 & 67.85 & 107.27 & 55.77 \\
         GFP-GAN \cite{GFPGAN}          & 64.14 & \textcolor{blue}{\textit{59.20}} & 111.79 & 54.84 \\
         CodeFormer \cite{CF}       & 66.84 & 60.10 & 114.34 & 56.15 \\
        \midrule
         SPARNetHD \cite{SPAR}        & 69.79 & 61.34 & 110.30 & 52.28 \\
         VQFR \cite{VQFR}             & 81.37 & 60.84 & \textcolor{blue}{\textit{104.39}} & \textcolor{blue}{\textit{51.81}} \\
        \midrule
         \textbf{DR2 + SPAR(ours)} & \textcolor{blue}{\textit{61.66}} & 63.69 & 107.00 & 52.27 \\
         \textbf{DR2 + VQFR(ours)} & \textcolor{red}{\textbf{60.06}} & \textcolor{red}{\textbf{58.78}} & \textcolor{red}{\textbf{103.91}} & \textcolor{red}{\textbf{50.98}} \\
        \bottomrule
      \end{tabular}
    }
  \caption{Quantitative comparisons on \textbf{WIDER-Critical} (W-Cr), \textbf{WIDER-Normal} (W-Nm), \textbf{CelebChild} (Celeb-C) and \textbf{LFW-Test} (LFW). \textcolor{red}{\textbf{Red}} and \textcolor{blue}{\textit{blue}} indicates the best and the second best performance.}
  \label{Tab:2}
\end{table}

\noindent \textbf{Real-World Datasets.} We evaluate FID scores on different real-world datasets and present quantitative results in \cref{Tab:2}. On severely degraded dataset WIDER-Critical, our DR2 + VQFR and DR2 + SPAR achieve the best and the second best FID. On other datasets with only mild degradation, the restoration quality rather than robustness becomes the bottleneck, so DR2 + SPAR with no facial priors struggles to stand out, while DR2 + VQFR still achieves the best performance.

\begin{figure*}
\centering
\includegraphics[width=1.0\linewidth, trim=0 18 0 5 ]{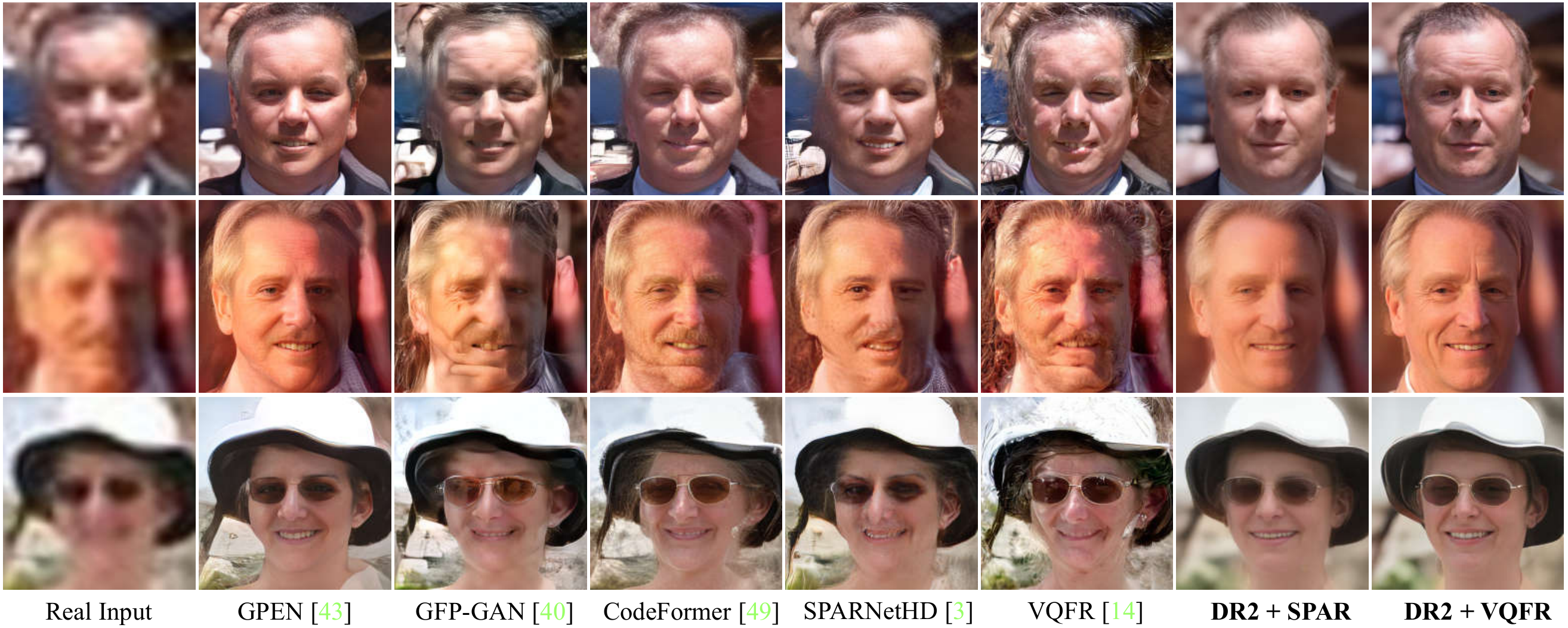}
\caption{Qualitative comparisons on \textbf{WIDER-critical} dataset. Thanks to the generative ability of DR2, our methods produce more visually pleasant results when inputs are of low-resolution.}
\label{Fig:5}
\end{figure*}

Qualitative results on WIDER-Critical are shown in \cref{Fig:5}. When input images' resolutions are very low, previous methods fail to complement adequate information for pleasant faces, while our outputs are visually more pleasant thanks to the generative ability of DDPM.

%============================================================================================
%====================== 4.3 Comparisons with Diffusion-based Methods ========================
%============================================================================================
\subsection{Comparisons with Diffusion-based Methods}

\begin{figure}
    \centering
    \includegraphics[width=1.0\columnwidth, trim=0 30 0 0]{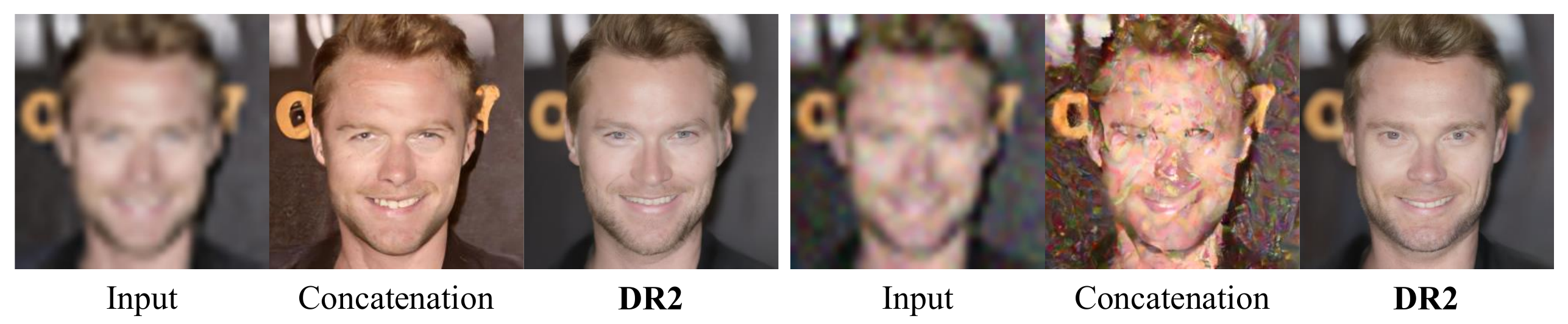}
    \caption{Comparisons with diffusion-based super-resolution method based on concatenation of input images. This method is highly degradation-sensitive.}
    \label{Fig:6}
\end{figure}

\begin{figure}
    \centering
    \includegraphics[width=1.0\columnwidth, trim=0 18 0 10]{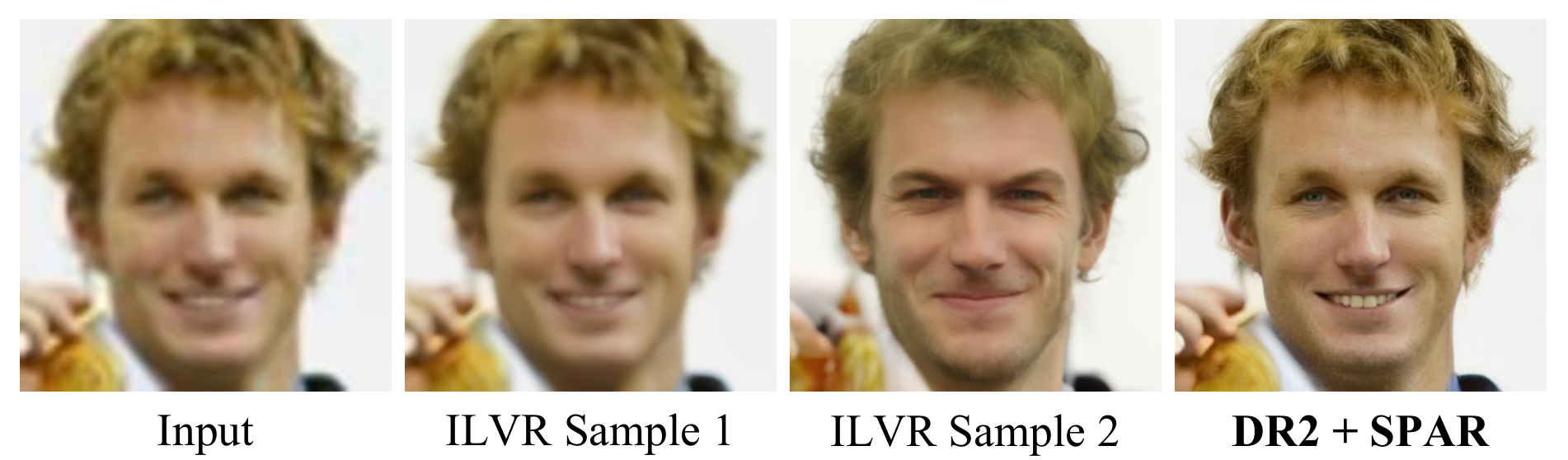}
    \caption{Trade-off problem of ILVR for blind face restoration. ILVR Sample 1 is sampled with wide conditioning step range \cite{ILVR} and Sample 2 with narrow conditioning range.}
    \label{Fig:7}
\end{figure}

Diffusion-based super-resolution methods can be grouped into two categories by whether feeding auxiliary input to the denoising U-Net. 

SR3 \cite{SR3_} typically uses the concatenation of low-resolution images and $\rvx_t$ as the input of the denoising U-Net. But SR3 fixes degradation to bicubic downsampling during training, which makes it highly degradation-sensitive. For visual comparisons, we re-implement the concatenation-based method based on \cite{DBG}. As shown in \cref{Fig:6}, minor noise in the second input evidently harm the performance of this concatenation-based method. Eventually, this type of method would rely on synthetic degradation to improve robustness like \cite{LDM}, while our DR2 have good robustness against different degradation without training on specifically degraded data.

Another category of methods is training-free, exploiting pretrained diffusion methods like ILVR \cite{ILVR}. It shows the ability to transform both clean and degraded low-resolution images into high-resolution outputs. However, relying solely on ILVR for blind face restoration faces the trade-off problem between fidelity and quality (realness). As shown in \cref{Fig:7}, ILVR Sample 1 has high fidelity to input but low visual quality because the conditioning information is over-used. On the contrary, under-use of conditions leads to high quality but low fidelity as ILVR Sample 2. In our framework, fidelity is controlled by DR2 and high-quality details are restored by the enhancement module, thus alleviating the trade-off problem.

% Diffusion-based FSR methods also face the common problem of slow sampling speed. They perform $T$ denoising steps to restore an image. Our proposed DR2 is relatively faster because it only performs about $0.25T$ (both can be reduced by recent works on faster DDPM sampling) denoising steps.

% %============================================================================================
% %=================================== 4.4 User Controllability ===============================
% %============================================================================================
\subsection{Effect of Different $N$ and $\tau$}
\label{Sec:4.4}

% The low-quality information is mainly provided by iterative refinement during DR2 denoising process. Downsampling factor $N$ and output step $\tau$ not only determine the semantic similarity of DR2 output to the input image, but also pivotal to whether $\hat{x}_0$ is "clean" for the following enhancement. 
In this section, we explore the property of DR2 output in terms of the controlling parameter $(N, \tau)$ so that we can have a better intuitions for choosing appropriate parameters for variant input data. To avoid the influence of the enhancement modules varying in structures, embedded facial priors, and training strategies, we only evaluate DR2 outputs with no enhancement. 

\begin{figure*}
    \centering
    \includegraphics[width=1.0\textwidth, trim=0 10 0 10]{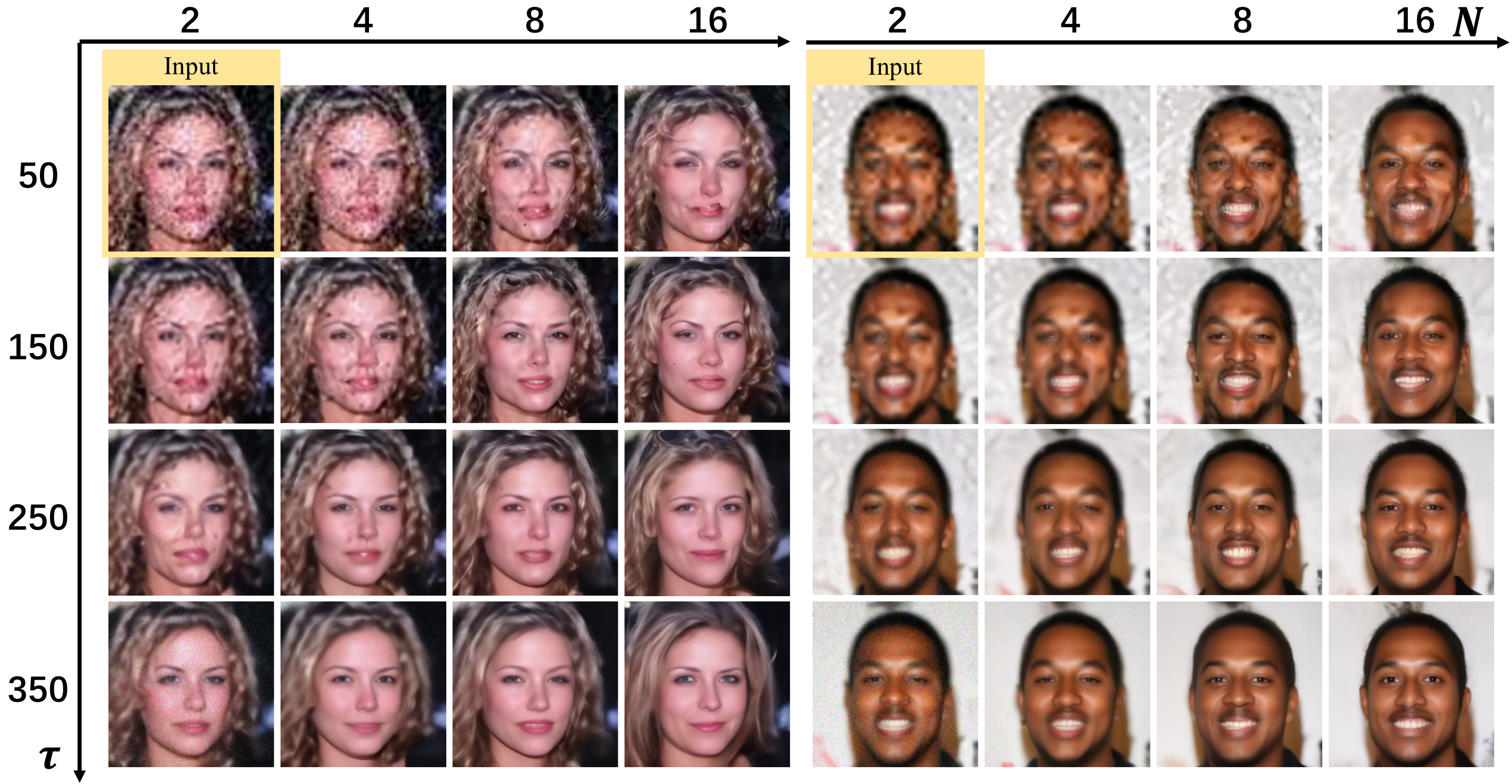}
    \caption{\textbf{DR2 outputs with various controlling parameters.} Input images are denoted with yellow boxes. Samples obtained from bigger $N$ and $\tau$ have lower fidelity but contain fewer degradation. Whiles samples from smaller $N$ and $\tau$ are more similar to the input but still contain degradation.}
    \label{Fig:8}
\end{figure*}

In \cref{Fig:8}, DR2 outputs are generated with different combinations of $N$ and $\tau$. Bigger $N$ and $\tau$ are effective to remove degradation but tent to make results deviant from the input. On the contrary, small $N$ and $\tau$ lead to high fidelity, but may keep the degradation in outputs.

\begin{figure}
    \centering
    \includegraphics[width=1.0\columnwidth, trim=0 10 0 20]{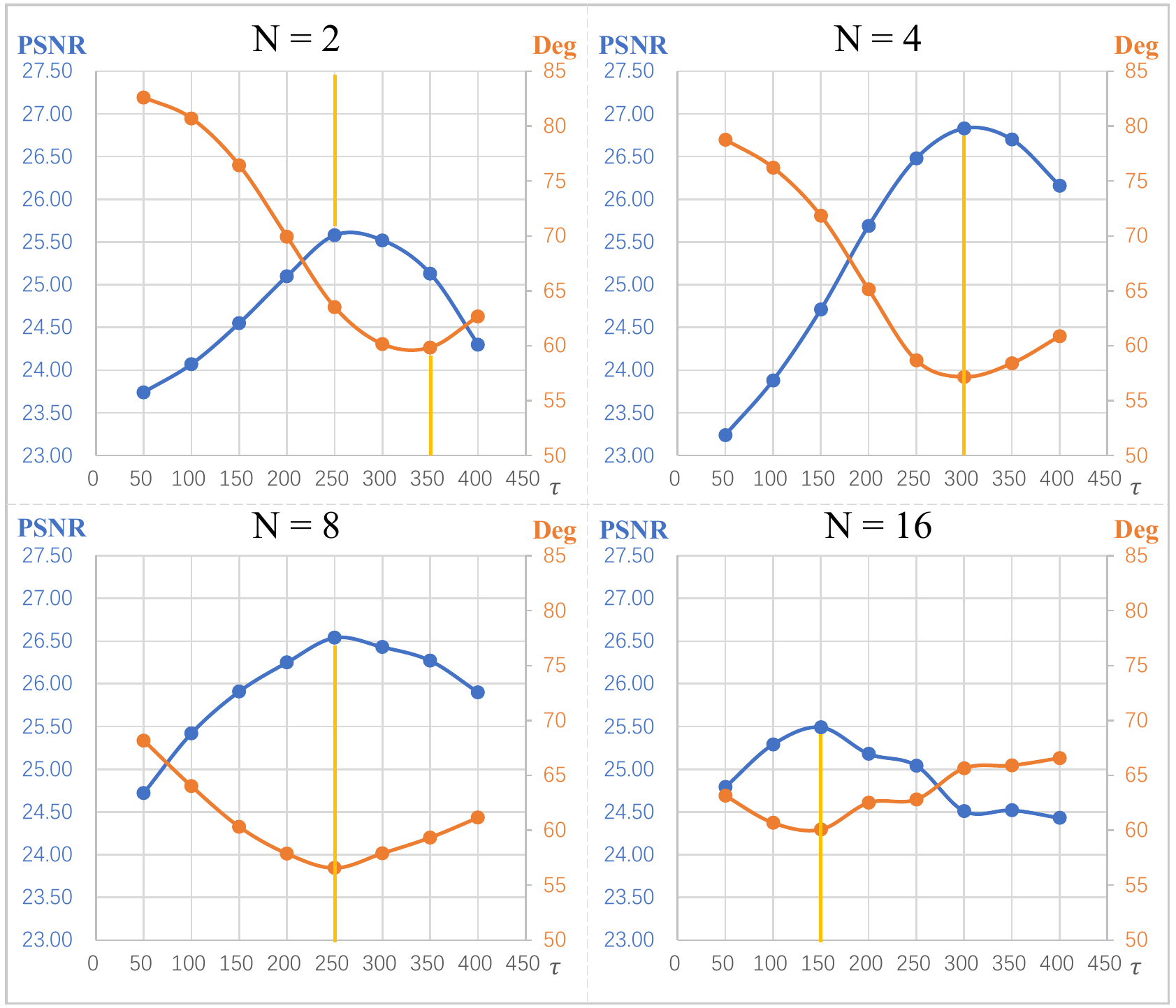}
    \caption{\textbf{Quantitative evaluation of DR2 output with various controlling parameters.} On different settings, both PSNR$\uparrow$ and Deg$\downarrow$ first get better then worse as $\tau$ increase. The \textcolor[RGB]{255,192,0}{golden} line indicates the best result for each metric.}
    \label{Fig:9}
\end{figure}

We provide quantitative evaluations on CelebA-Test ($8\times$, medium split) dataset in \cref{Fig:9}. With bicubically downsampled low-resolution images used as ground truth, we adopt pixel-wise metric (PSNR$\uparrow$) and identity distance (Deg$\downarrow$) based on the embedding angle of ArcFace \cite{GFPGAN10} for evaluating the quality and fidelity of DR2 outputs. For scale $N$ = 4, 8, and 16, PSNR first goes up and Deg goes down because degradation is gradually removed as $\tau$ increases. Then they hit the optimal point at the same time before the outputs begin to deviate from the input as $\tau$ continues to grow. Optimal $\tau$ is bigger for smaller $N$. For $N = 2$, PSNR stops to increase before Deg reaches the optimality because Gaussian noise starts to appear in the output (like results sampled with $(N, \tau)=(2,350)$ in \cref{Fig:8}). This cause of the appearance of Gaussian noise is that $\rvy_t$ sampled by \cref{Eq:forw_detail} contains heavy Gaussian noise when $t$ $(t > \tau)$ is big and most part of $\rvy_t$ is utilized by \cref{Eq:ILVR} when $N$ is small.

% ============================================================================================
% ============================== 4.5 Discussion and Limitations ==============================
% ============================================================================================
\subsection{Discussion and Limitations}
Our DR2 is built on a pretrained DDPM, so it would face the problem of slow sampling speed even we only perform $0.25T$ steps in total. But DR2 can be combined with diffusion acceleration methods like inference every 10 steps. And keep the output resolution of DR2 relatively low ($256^2$ in our practice) and leave the upsampling for enhancement module for faster speed.

Another major limitation of our proposed DR2 is the manual choosing for controlling parameters $N$ and $\tau$.
% Luckily, we do not need to do a full range grid search. For the most cases, N = 4 or 8 would satisfy. As for τ , we first empirically choose an initial value between 250 and 50. Then, we can decrease or increase τ 20 by 20 or even 50 by 50 accordingly. The search space is not very large and users can learn to choose appropriate parameters quickly. 
As a future work, we are exploring whether image quality assessment scores (like NIQE) can be used to develop an automatic search algorithms for $N$ and $\tau$.

Furthermore, for inputs with slight degradation, DR2 is less necessary because previous methods can also be effective and faster. And in extreme cases where input images contains very slight degradation or even no degradation, DR2 transformation may remove details in the inputs, but that is not common cases for blind face restoration.

\section{Conclusion}
\label{sec:conc}

We propose the DR2E, a two-stage blind face restoration framework that leverages a pretrained DDPM to remove degradation from inputs, and an enhancement module for detail restoration. In the first stage, DR2 removes degradation by using diffused low-quality information as conditions to guide the generative process. This transformation requires no synthetically degraded data for training. Extensive comparisons demonstrate the strong robustness and high restoration quality of our DR2E framework.

\section*{Acknowledgements}
This work is supported by National Natural Science Foundation of China (62271308), Shanghai Key Laboratory
of Digital Media Processing and Transmissions (STCSM 22511105700, 18DZ2270700), 111 plan (BP0719010), and State Key Laboratory of UHD Video and Audio Production and Presentation.

%%%%%%%%% REFERENCES
{\small
\bibliographystyle{ieee_fullname}
\bibliography{main}
}

\appendix
\clearpage
\counterwithin{figure}{section}
\counterwithin{table}{section}
\renewcommand\thesection{\Alph{section}}
\renewcommand\thefigure{A\arabic{figure}}
\renewcommand\thetable{A\arabic{table}}

{\centering\section*{Appendix}}

In the appendix, we provide additional discussions and results complementing \cref{sec:exp}. In \cref{Sec:A}, we conduct further ablation studies on initial condition and iterative refinement to show the control and conditioning effect  these two mechanisms bring to the DR2 generative process. In \cref{Sec:B}, we provide (1) our detailed settings of DR2 controlling parameters $(N,\tau)$ for each testing dataset, and (2) show more qualitative comparisons on each split of CelebA-Test dataset in this section to illustrate how our methods and previous state-of-the-art methods perform over variant levels of degradation.

% ============================================================================================================
% =========================================== A. More Ablation Studies =======================================
% ============================================================================================================
\section{More Ablation Studies}
\label{Sec:A}
In this section, we explore the effect of initial condition and iterative refinement in DR2. To avoid the influence of the enhancement modules varying in structures, embedded facial priors, and training strategies, we only conduct experiments on DR2 outputs with no enhancement. To evaluate the degradation removal performance and fidelity of DR2 outputs, we use bicubic downsampled images as ground truth low-resolution (GT LR) image. This is intuitive as DR2 is targeted to produce clean but blurry middle results.

% ------------------------------------------------------------------------------------------------------------
% --------------------------------------- A. 1 Effect of Initial Condition------------------------------------
% ------------------------------------------------------------------------------------------------------------
\subsection{Conditioning Effect of Initial Condition with Iterative Refinement Enabled}

% ------------------------------------------------------------------------------------------------------------
% ----------------------------------------------- A. 1 . 1 ---------------------------------------------------
% ------------------------------------------------------------------------------------------------------------
% \subsubsection*{With Iterative Refinement Enabled}
% \noindent (1) \textbf{With Iterative Refinement Enabled}

During DR2 generative process, diffused low-quality inputs is provided through initial condition and iterative refinement. The latter one yields stronger control to the generative process because it is performed at each step, while initial condition only provides information in the beginning with heavy Gaussian noise attached. To quantitatively evaluate the effect of initial condition, we follow the settings of S\cref{Sec:4.4} by calculating the pixel-wise metric (PSNR) and identity distance (Deg) between DR2 outputs and ground truth low-resolution images on CelebA-Test ($8\times$, medium split) dataset. Quantitative results are shown in \cref{Tab:A1}. We fix $(N, \tau) = (4, 300)$ and change the value of $\omega$. When $\omega = 1000 = T$, no initial condition is provided because $\mathbf{y}_{1000}$ is pure Gaussian noise. As shown in the table, with iterative refinement providing strong control to DR2 generative process, the quality and fidelity of DR2 outputs are not evidently affected as $\omega$ varies. 

\begin{figure}[h]
\centering
\includegraphics[width=1.0\linewidth, trim=0 15 0 10 ]{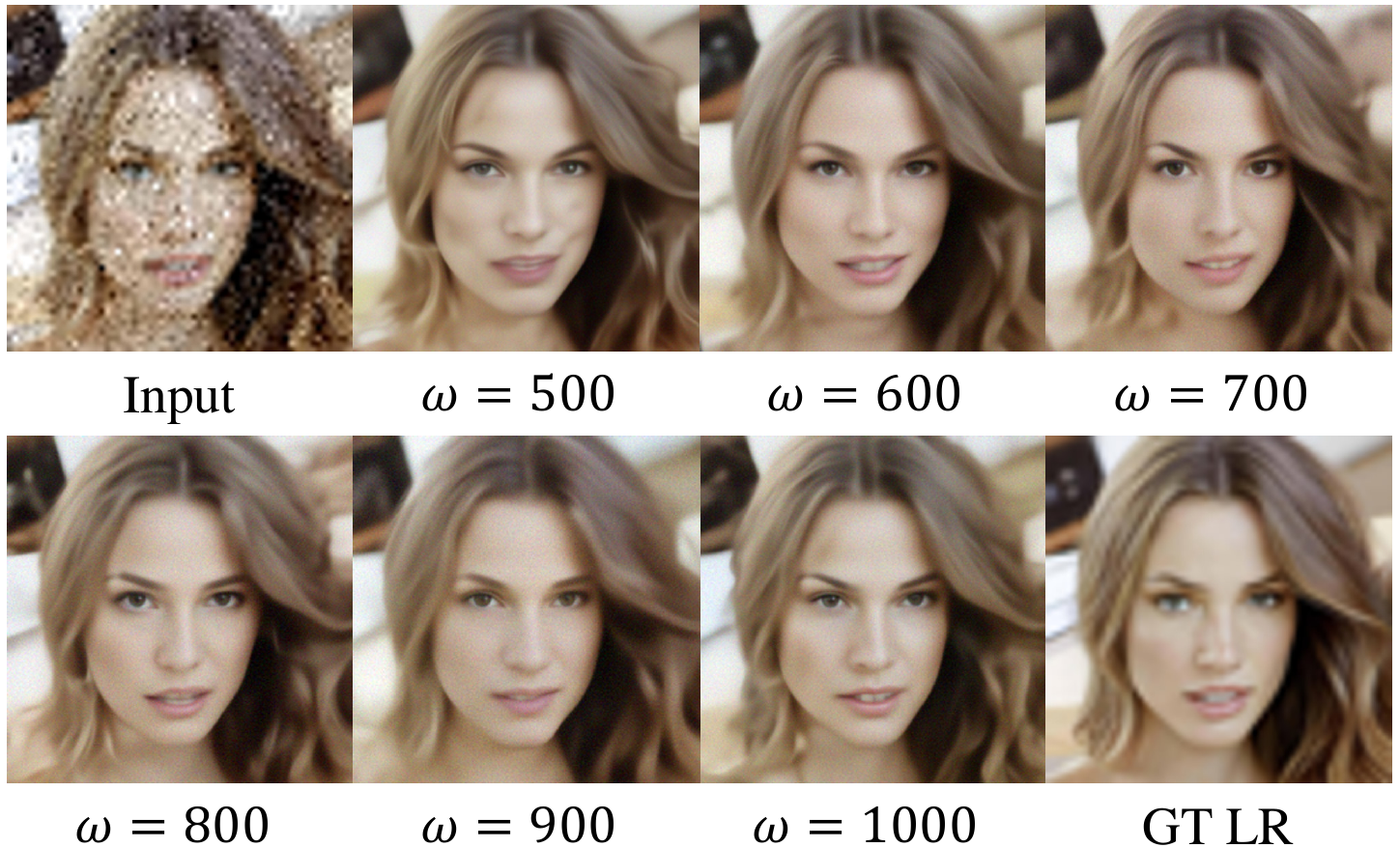}
\caption{\textbf{Qualitative effect of $\omega$ with iterative refinement enabled.} Testing image is from CelebA-Test ($8\times$, medium split) dataset with iterative refinement controlling parameters set as $(N, \tau) = (4, 300)$.}
\label{Fig:A1}
\end{figure}

Qualitative results are provided in \cref{Fig:A1}. With fixed iterative refinement controlling parameters, $\omega$ has little visual effect on DR2 outputs. Although the initial condition provides limited information compared with iterative refinement, it significantly reduces the total steps of DR2 denoising process. 

\begin{table}[h]
  \centering
  \resizebox{1.0\columnwidth}{!}{
      \begin{tabular}{c|c c c c c c c c}
        \toprule
        $\omega$ & 350 & 400 & 500 & 600 & 700 & 800 & 900 & 1000 \\
        \midrule
         PSNR$\uparrow$     & 26.86 & 26.87 & 26.87 & 26.83 & 26.80 & 26.77 & 26.77 & 26.76 \\
         Deg $\downarrow$   & 56.15 & 56.02 & 56.03 & 56.94 & 57.38 & 57.86 & 57.67 & 57.78 \\ 
        \bottomrule
      \end{tabular}
    }
  \caption{\textbf{Effect of different $\omega$ with iterative refinement enabled.} With iterative refinement, initial condition has little effect on DR2 output quality as long as $\omega - \tau$ is not two small.}
  \label{Tab:A1}
\end{table}

\begin{table}[h]
  \centering
  \resizebox{1.0\columnwidth}{!}{
      \begin{tabular}{c|c|c c c c c}
        \toprule
        $\tau$ & $\omega$ & 300 & 400 & 500 & 600 & 700 \\
        \midrule
        \multirow{2}*{300} & PSNR$\uparrow$   & - & 24.79 & 22.95 & 20.84 & 18.34 \\
        ~                  & Deg$\downarrow$  & - & 61.87 & 68.53 & 74.84 & 79.84 \\
        \midrule
        \multirow{2}*{150} & PSNR$\uparrow$   & 24.58 & 24.05 & 22.57 & 20.61 & 18.22 \\
        ~                  & Deg$\downarrow$  & 63.07 & 64.24 & 69.64 & 75.04 & 80.51 \\
        \midrule
        \multirow{2}*{0} & PSNR$\uparrow$   & 23.98 & 23.70 & 22.33 & 20.47 & 18.17 \\
        ~                  & Deg$\downarrow$  & 64.70 & 64.29 & 69.71 & 74.83 & 80.50 \\
        \bottomrule
      \end{tabular}
    }
  \caption{\textbf{Quantitative results with iterative refinement disabled.} Without iterative refinement, initial condition can only provide limited control to the generative process especially when $\omega$ is big.}
  \label{Tab:A2}
\end{table}

% ------------------------------------------------------------------------------------------------------------
% ----------------------------------------------- A. 1 . 2 ---------------------------------------------------
% ------------------------------------------------------------------------------------------------------------
\begin{figure*}[ht!]
\centering
\includegraphics[width=1.0\linewidth, trim=0 8 0 10 ]{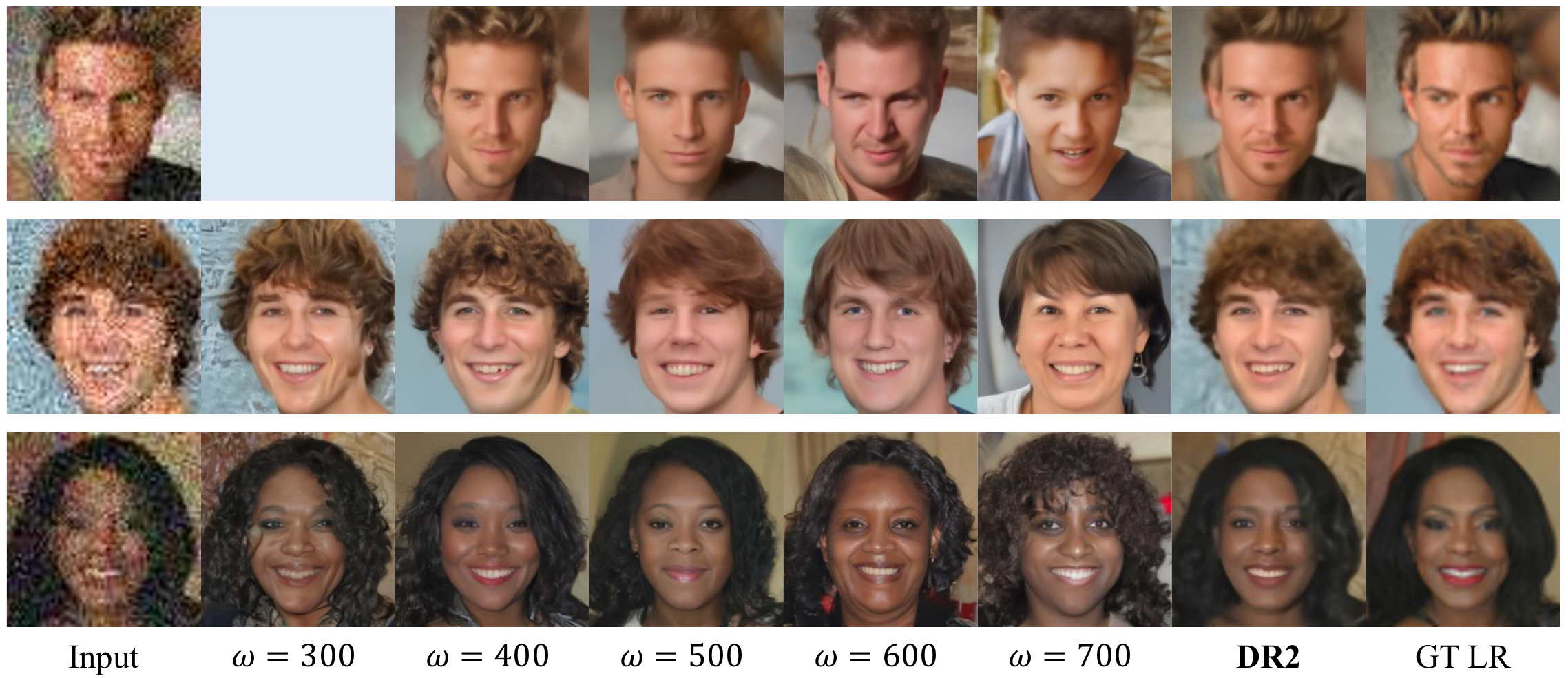}
\caption{\textbf{Qualitative results without iterative refinement.} Testing images are from CelebA-Test ($8\times$, medium split) dataset. The three rows are sampled with $\tau = 300$, 150, and 0 respectively. When $\tau = 0$, truncated output is no longer needed. DR2 outputs are sampled with iterative refinement controlling parameters set as $(N, \tau) = (4, 300)$.}
\label{Fig:A2}
\end{figure*}

\begin{table*}[h!]
  \centering
      \begin{tabular}{c|c|c c|c c|c c|c c|c c|c c|c c}
        \toprule
        ~ & ~ & \multicolumn{2}{c|}{CA $\times16$} & \multicolumn{2}{c|}{CA $\times8$} & \multicolumn{2}{c|}{CA $\times4$} & \multicolumn{2}{c|}{W-Cr} & \multicolumn{2}{c|}{W-Nm} & \multicolumn{2}{c|}{CelebC} & \multicolumn{2}{c}{LFW}\\
         Methods & Split & $N$ & $\tau$ & $N$ & $\tau$ & $N$ & $\tau$ & $N$ & $\tau$  & $N$ & $\tau$  & $N$ & $\tau$  & $N$ & $\tau$ \\
        \midrule
        \multirow{3}*{DR2 + SPAR} & mild   &  8 & 220 & 4 & 200 & 4 & 80  & \multirow{3}*{16} & \multirow{3}*{200} & \multirow{3}*{8} & \multirow{3}*{100} & \multirow{3}*{4} & \multirow{3}*{10} & \multirow{3}*{4} & \multirow{3}*{60}\\
        ~                         & medium & 16 & 320 & 8 & 350 & 4 & 220 & & & & & & & \\
        ~                         & severe & 32 & 250 & 8 & 370 & 8 & 190 & & & & & & & \\
        \midrule
        \multirow{3}*{DR2 + VQFR} & mild   &  8 & 250 & 4 & 200 & 4 & 80  & \multirow{3}*{8} & \multirow{3}*{250} & \multirow{3}*{8} & \multirow{3}*{100} & \multirow{3}*{4} & \multirow{3}*{30} & \multirow{3}*{4} & \multirow{3}*{60}\\
        ~                         & medium & 16 & 300 & 4 & 350 & 4 & 180 & & & & & & &\\
        ~                         & severe & 32 & 250 & 8 & 370 & 8 & 190 & & & & & & &\\
        \bottomrule
      \end{tabular}
  \caption{\textbf{Controlling parameter settings} for \textbf{CelebA-Test} (CA), \textbf{WIDER-Critical} (W-Cr), \textbf{WIDER-Normal} (W-Nm), \textbf{CelebChild} (CelebC) and \textbf{LFW-Test} (LFW). For more severely degraded dataset, bigger $N$ and $\tau$ are adopted and vice versa.}
  \label{Tab:A3}
\end{table*}

% \subsubsection{With Iterative Refinement Disabled}
\subsection{Conditioning Effect of Initial Condition with Iterative Refinement Disabled}

We conduct experiments without iterative refinement in this section to show that generative results bear less fidelity to the input without it. Without iterative refinement, DR2 generative process relies solely on the initial condition to utilize information of low-quality inputs, and generate images through DDPM denoising steps stochastically from initial condition. $\omega$ now becomes an important controlling parameter determining how much conditioning information is provided. We also calculate PSNR and Deg between DR2 outputs and ground-truth low-resolution images on  CelebA-Test ($8\times$, medium split) dataset. Quantitative results with different $(\omega, \tau)$ are provided in \cref{Tab:A2}. Note that PSNR and Deg are all worse than those in \cref{Tab:A1}, and have a negative correlation with $\omega$ because less information of inputs is used as $\omega$ increases.

Qualitative results are shown in \cref{Fig:A2}. When $\omega \geqslant 400$, added noise in initial condition is strong enough to cover the degradation in inputs so the output tends to be smooth and clean. But as $\omega$ increases, the outputs become more irrelevant to the input because the initial conditions are weakened. Compared with results that were sampled with iterative refinement, the importance of it on preserving semantic information is obvious.

% ============================================================================================================
% =========================================== B. Detailed Comparisons =======================================
% ============================================================================================================
\section{Detailed Settings and Comparisons}
\label{Sec:B}

\subsection{Controlling Parameter Settings}
As introduced in \cref{sec:4.1}, to evaluate the performance on different levels of degradation, we synthesize three splits (mild, medium, and severe) for each upsampling task ($16\times$, $8\times$, and $4\times$) together with four real-world datasets. During the experiment in \cref{sec:4.2}, different controlling parameters $(N, \tau)$ are used for each dataset or split. Generally speaking, big $N$ and $\tau$ are more effective to remove the degradation but lead to lower fidelity and vice versa. We provide detailed settings we employed in \cref{Tab:A3}

\subsection{More Qualitative Comparisons}
\begin{figure*}[h]
\centering
\includegraphics[width=1.0\linewidth, trim=0 8 0 10 ]{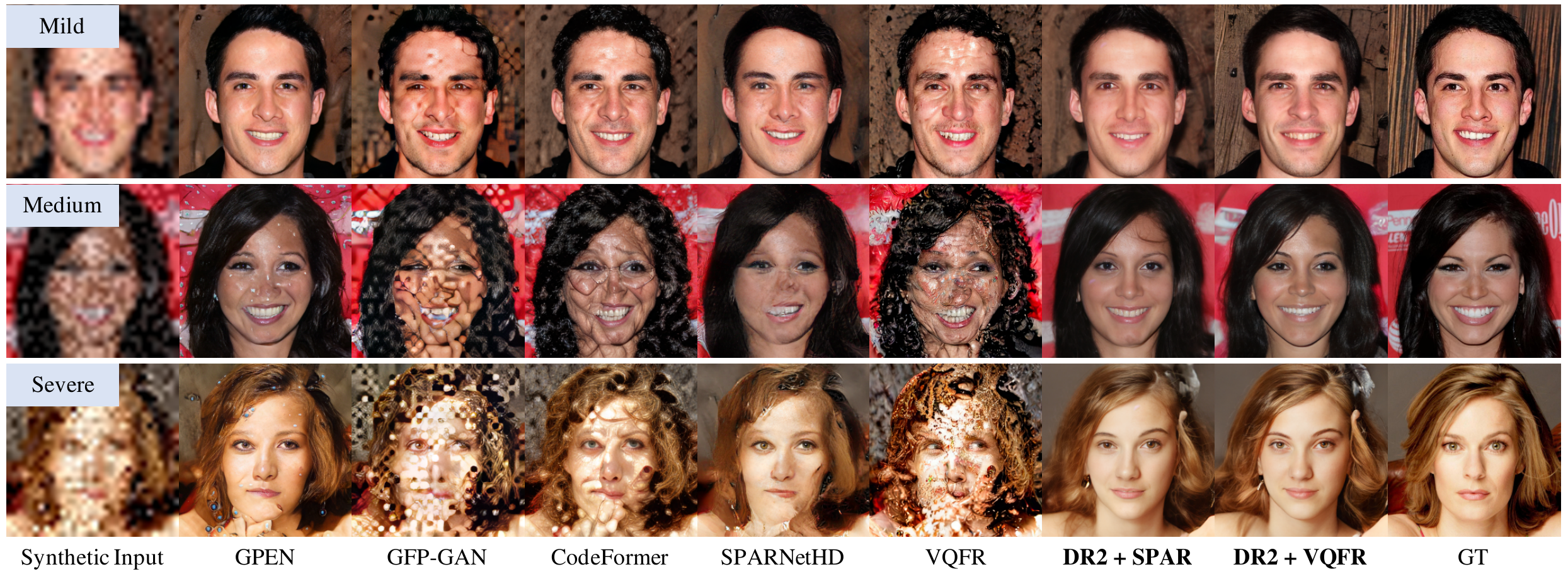}
\caption{\textbf{Qualitative results on CelebA-Test ($\times16$).} Previous methods produce more artifacts when inputs are heavily degraded.}
\label{Fig:A3}
\end{figure*}

\begin{figure*}[h]
\centering
\includegraphics[width=1.0\linewidth, trim=0 8 0 10 ]{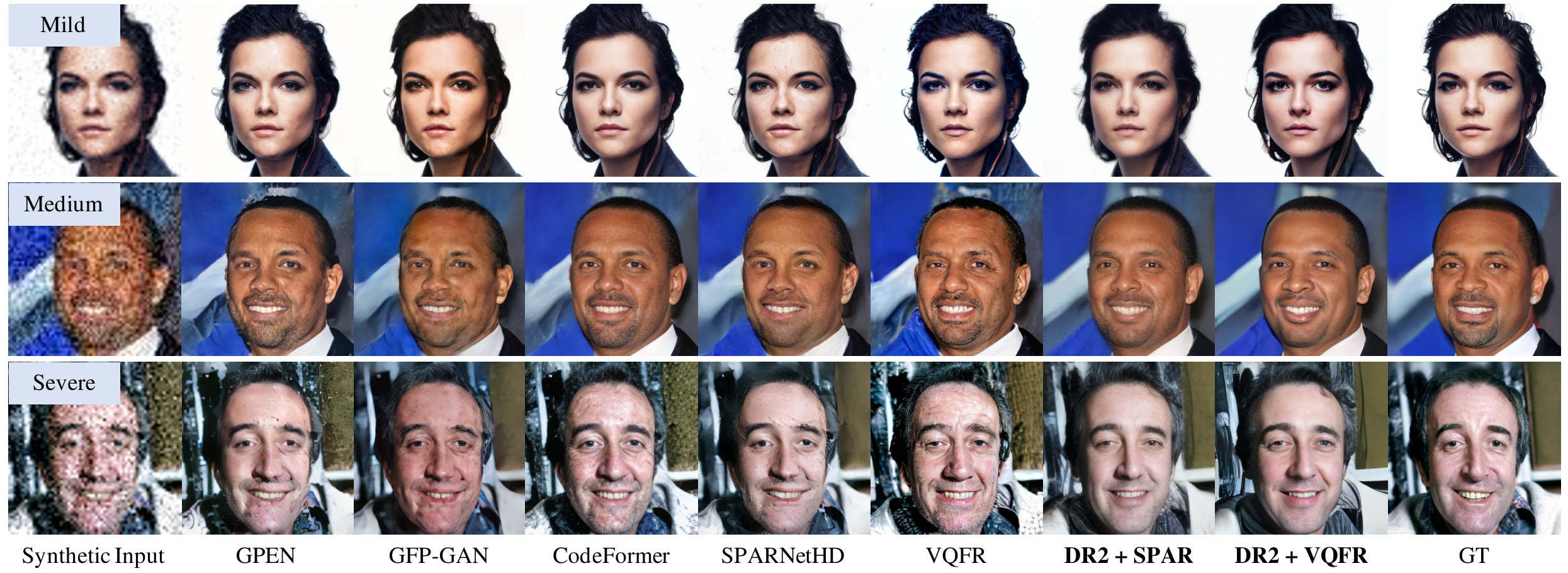}
\caption{\textbf{Qualitative results on CelebA-Test ($\times8$).} Previous methods produce more artifacts when inputs are heavily degraded.}
\label{Fig:A4}
\end{figure*}

\begin{figure*}[h]
\centering
\includegraphics[width=1.0\linewidth, trim=0 8 0 10 ]{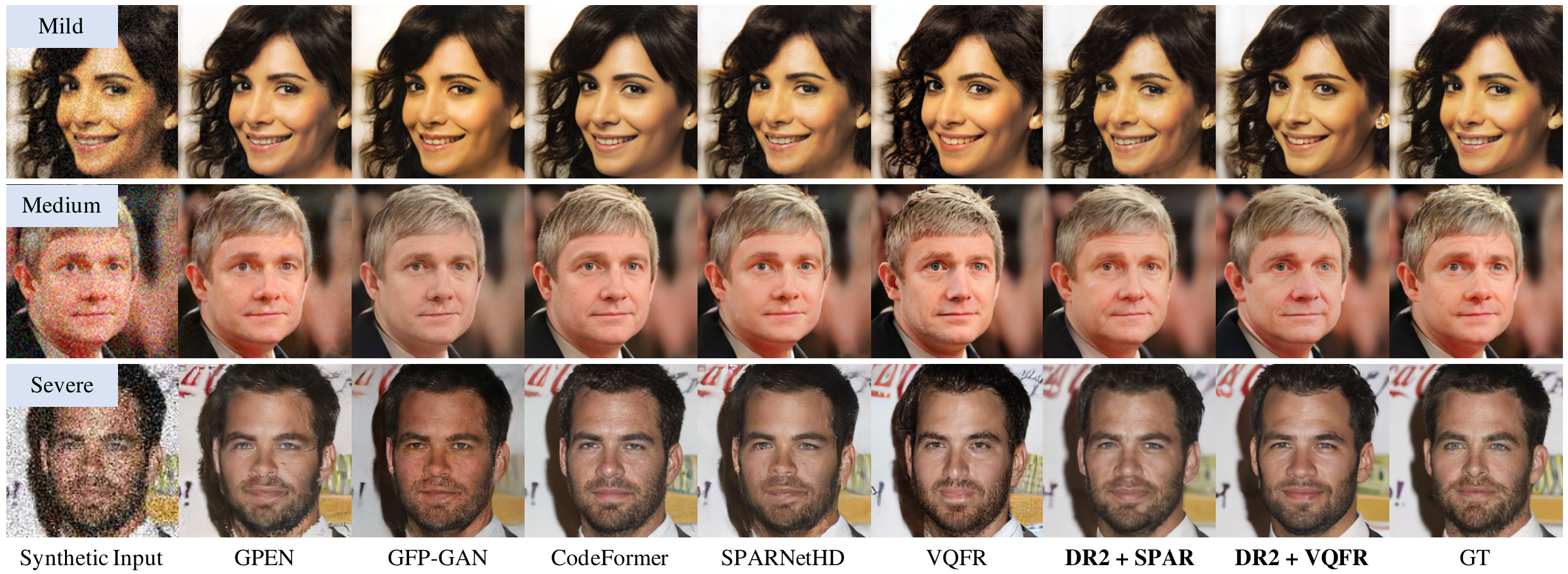}
\caption{\textbf{Qualitative results on CelebA-Test ($\times4$).} Our methods produce comparable results with previous arts on mildly degraded data.}
\label{Fig:A5}
\end{figure*}

For more comprehensive comparisons with previous methods on different levels of degraded dataset, we provide qualitative results on each split of CelebA-Test dataset under each upsampling factor in \cref{Fig:A3,Fig:A4,Fig:A5}. As shown in the figures, for inputs with slight degradation, DR2 transformation is less necessary because previous methods can also be effective. But for severe degradation, previous methods fail since they never see such degradation during training. While our method shows great robustness even though no synthetic degraded images are employed for training.

\end{document}

{\centering\section*{Supplementary}}
% \title{Appendix}  % **** Enter the paper title here

% \maketitle
% \thispagestyle{empty}
% \appendix
\renewcommand\thesection{\Alph{section}}
\renewcommand\thefigure{S\arabic{figure}}
\renewcommand\thetable{S\arabic{table}}

In the appendix, we provide additional discussions and results based on Section \textcolor{red}{4}. In \cref{Sec:A}, we conduct further ablation studies on initial condition and iterative refinement. In \cref{Sec:B}, we provide (1)our detailed settings of DR2 controlling parameters $(N,\tau)$ for each testing dataset, and (2) present full quantitative comparisons with previous methods on each split of CelebA-Test dataset. We also (3) show more qualitative comparisons on each split of CelebA-Test dataset in this section.

% ============================================================================================================
% =========================================== A. More Ablation Studies =======================================
% ============================================================================================================
\section{More Ablation Studies}
\label{Sec:A}
In this section, we explore the effect of initial condition and iterative refinement in DR2. To avoid the influence of the enhancement modules varying in structures, embedded facial priors, and training strategies, we only conduct experiments on DR2 outputs with no enhancement. 
% ------------------------------------------------------------------------------------------------------------
% --------------------------------------- A. 1 Effect of Initial Condition------------------------------------
% ------------------------------------------------------------------------------------------------------------
\subsection{Effect of Initial Condition $\mathbf{y}_\omega$ with Iterative Refinement Enabled}
During DR2 generative process, diffused low-quality inputs is provided through initial condition and iterative refinement. The latter one intuitively yields stronger control to the generative process because it is performed at each step, while initial condition only provides information in the beginning with heavy Gaussian noise. To quantitatively evaluate the effect of initial condition, we follow the settings of Section \textcolor{red}{4.4} by calculating the pixel-wise metric (PSNR) and identity distance (Deg) between DR2 outputs and ground truth low-resolution images on  CelebA-Test ($8\times$, medium split) dataset. Quantitative results are shown in \cref{Tab:A1}. We fix $(N, \tau) = (4, 300)$ and change the value of $\omega$. When $\omega = 1000 = T$, no initial condition is provided because $\mathbf{y}_{1000}$ is pure Gaussian noise, but the quality and fidelity of DR2 outputs are not evidently harmed at this point. 

\begin{figure}[h]
\centering
\includegraphics[width=1.0\linewidth, trim=0 15 0 10 ]{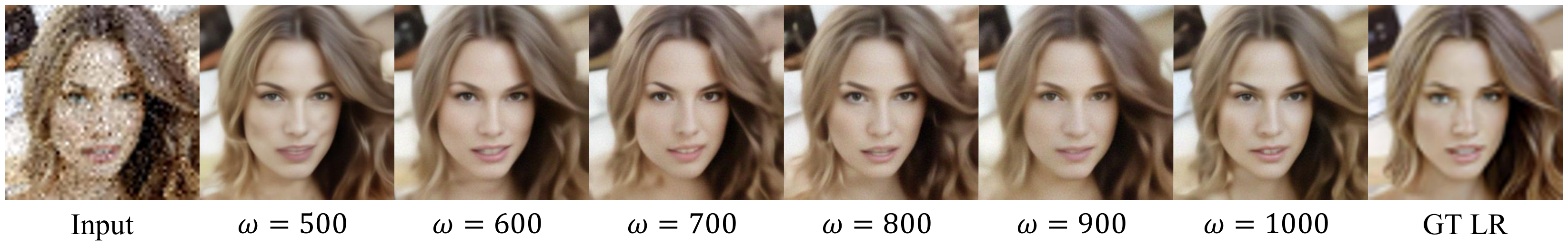}
\caption{\textbf{Qualitative effect of $\omega$.} Testing image is from CelebA-Test ($8\times$, medium split) dataset with iterative refinement controlling parameters set as $(N, \tau) = (4, 300)$.}
\label{Fig:A1}
\end{figure}

Qualitative results are provided in \cref{Fig:A1}. With fixed iterative refinement controlling parameters, $\omega$ has little visual effect on DR2 outputs. Although the initial condition provides limited information compared with iterative refinement, it significantly reduces the total steps of DR2 denoising process. 

\begin{table}[h]
  \centering
  % \resizebox{\textwidth}{5mm}{
      \begin{tabular}{c|c c c c c c c c}
        \toprule
        $\omega$ & 350 & 400 & 500 & 600 & 700 & 800 & 900 & 1000 \\
        \midrule
         PSNR$\uparrow$     & 26.86 & 26.87 & 26.87 & 26.83 & 26.80 & 26.77 & 26.77 & 26.76 \\
         Deg $\downarrow$   & 56.15 & 56.02 & 56.03 & 56.94 & 57.38 & 57.86 & 57.67 & 57.78 \\ 
        \bottomrule
      \end{tabular}
    % }
  \caption{\textbf{Effect of $\omega$.} With iterative refinement, initial condition has little effect on DR2 output quality as long as $\omega - \tau$ is not two small.}
  \label{Tab:A1}
\end{table}

% ------------------------------------------------------------------------------------------------------------
% -------------------------------- A. 2 Importance of Iterative Refinement------------------------------------
% ------------------------------------------------------------------------------------------------------------
\subsection{Effect of Initial Condition $\mathbf{y}_\omega$ with Iterative Refinement Disabled}
We conduct experiments without iterative refinement in this section to show that generative results are not close to the input without it. Without iterative refinement, DR2 generative process relies solely on the initial condition to utilize information of low-quality inputs, and generate images through DDPM denoising steps stochastically from initial condition. So $\omega$ becomes an important controlling parameter while downsampling factor $N$ is no longer needed. We also calculate PSNR and Deg between DR2 outputs and ground-truth low-resolution images on  CelebA-Test ($8\times$, medium split) dataset. Quantitative results with different $(\omega, \tau)$ are provided in \cref{Tab:A2}. Note that PSNR and Deg are all worse than those in \cref{Tab:A1}, and have a negative correlation with $\omega$ because less information of inputs is used as $\omega$ increases. 

Qualitative results are shown in \cref{Fig:A2}. When $\omega \geqslant 400$, added noise in initial condition is strong enough to cover the degradation in inputs so the output tends to be smooth and clean. But as $\omega$ increases, the outputs become more irrelevant to the input because the initial conditions are weakened. Compared with results that were sampled with iterative refinement, the importance of it on preserving semantic information is obvious.

\begin{table}[h]
  \centering
  % \resizebox{\columnwidth}{20mm}{
      \begin{tabular}{c|c|c c c c c}
        \toprule
        $\tau$ & $\omega$ & 300 & 400 & 500 & 600 & 700 \\
        \midrule
        \multirow{2}*{300} & PSNR$\uparrow$   & - & 24.79 & 22.95 & 20.84 & 18.34 \\
        ~                  & Deg$\downarrow$  & - & 61.87 & 68.53 & 74.84 & 79.84 \\
        \midrule
        \multirow{2}*{150} & PSNR$\uparrow$   & 24.58 & 24.05 & 22.57 & 20.61 & 18.22 \\
        ~                  & Deg$\downarrow$  & 63.07 & 64.24 & 69.64 & 75.04 & 80.51 \\
        \midrule
        \multirow{2}*{0} & PSNR$\uparrow$   & 23.98 & 23.70 & 22.33 & 20.47 & 18.17 \\
        ~                  & Deg$\downarrow$  & 64.70 & 64.29 & 69.71 & 74.83 & 80.50 \\
        \bottomrule
      \end{tabular}
    % }
  \caption{\textbf{Quantitative results without iterative refinement.} Without iterative refinement, initial condition can only provide limited control to the generative process especially when $\omega$ is big.}
  \label{Tab:A2}
\end{table}

\begin{figure}[h]
\centering
\includegraphics[width=1.0\linewidth, trim=0 8 0 10 ]{cvpr2023-author_kit-v1_1-1/latex/figs/figsup2_effect_of_w_without_IR.pdf}
\caption{\textbf{Qualitative results without iterative refinement.} Testing images are from CelebA-Test ($8\times$, medium split) dataset. The three rows are sampled with $\tau = 300$, 150, and 0 respectively. When $\tau = 0$, truncated output is no longer needed. DR2 outputs are sampled with iterative refinement controlling parameters set as $(N, \tau) = (4, 300)$.}
\label{Fig:A2}
\end{figure}

% ============================================================================================================
% =========================================== B. Detailed Comparisons =======================================
% ============================================================================================================
\section{Detailed Settings and Comparisons}
\label{Sec:B}

\subsection{Controlling Parameter Settings}
As introduced in Section \textcolor{red}{4.1}, to evaluate the performance on different levels of degradation, we synthesize three splits (mild, medium, and severe) for each upsampling task ($16\times$, $8\times$, and $4\times$) together with four real-world datasets. During the experiment in Section \textcolor{red}{4.2}, different controlling parameters $(N, \tau)$ are used for each dataset or split. Generally speaking, big $N$ and $\tau$ are more effective to remove the degradation but lead to lower fidelity and vice versa. We provide detailed settings in \cref{Tab:A3}

\begin{table}
  \centering
      \begin{tabular}{c|c|c c|c c|c c|c c|c c|c c|c c}
        \toprule
        ~ & ~ & \multicolumn{2}{c|}{CA $\times16$} & \multicolumn{2}{c|}{CA $\times8$} & \multicolumn{2}{c|}{CA $\times4$} & \multicolumn{2}{c|}{W-Cr} & \multicolumn{2}{c|}{W-Nm} & \multicolumn{2}{c|}{CelebC} & \multicolumn{2}{c}{LFW}\\
         Methods & Split & $N$ & $\tau$ & $N$ & $\tau$ & $N$ & $\tau$ & $N$ & $\tau$  & $N$ & $\tau$  & $N$ & $\tau$  & $N$ & $\tau$ \\
        \midrule
        \multirow{3}*{DR2 + SPAR} & mild   &  8 & 220 & 4 & 200 & 4 & 80  & \multirow{3}*{16} & \multirow{3}*{200} & \multirow{3}*{8} & \multirow{3}*{10} & \multirow{3}*{4} & \multirow{3}*{10} & \multirow{3}*{4} & \multirow{3}*{60}\\
        ~                         & medium & 16 & 320 & 8 & 350 & 4 & 220 & & & & & & & \\
        ~                         & severe & 32 & 250 & 8 & 370 & 8 & 190 & & & & & & & \\
        \midrule
        \multirow{3}*{DR2 + VQFR} & mild   &  8 & 250 & 4 & 200 & 4 & 80  & \multirow{3}*{8} & \multirow{3}*{250} & \multirow{3}*{8} & \multirow{3}*{100} & \multirow{3}*{4} & \multirow{3}*{30} & \multirow{3}*{4} & \multirow{3}*{60}\\
        ~                         & medium & 16 & 300 & 4 & 350 & 4 & 180 & & & & & & &\\
        ~                         & severe & 32 & 250 & 8 & 370 & 8 & 190 & & & & & & &\\
        \bottomrule
      \end{tabular}
  \caption{Controlling parameter settings for \textbf{CelebA-Test} (CA), \textbf{WIDER-Critical} (W-Cr), \textbf{WIDER-Normal} (W-Nm), \textbf{CelebChild} (CelebC) and \textbf{LFW-Test} (LFW). For more severely degraded dataset, bigger $N$ and $\tau$ are adopted and vice versa.}
  \label{Tab:A3}
\end{table}

% \begin{table}
%   \centering
%       \begin{tabular}{c|c c|c c|c c|c c}
%         \toprule
%         ~ & \multicolumn{2}{c|}{W-Cr} & \multicolumn{2}{c|}{W-Nm} & \multicolumn{2}{c}{CelebC} & \multicolumn{2}{c}{LFW} \\
%          Methods & $N$ & $\tau$ & $N$ & $\tau$ & $N$ & $\tau$ & $N$ & $\tau$\\
%         \midrule
%         DR2 + SPAR  & 16 & 20 & 8 & 10 & 4 & 1 & 4 & 6  \\
%         \midrule
%         DR2 + VQFR  & 8 & 25 & 8 & 10 & 4 & 3 & 4 & 6  \\
%         \bottomrule
%       \end{tabular}
%   \caption{Controlling parameter settings for \textbf{WIDER-Critical} (W-Cr), \textbf{WIDER-Normal} (W-Nm), \textbf{CelebChild} (CelebC) and \textbf{LFW-Test} (LFW). For more severely degraded dataset WIDER-Critical, bigger $N$ and $\tau$ are adopted.}
%   \label{Tab:A4}
% \end{table}

\subsection{Detailed Comparisons with State-of-the-art Methods}

We calculate four evaluation metrics (LPIPS, FID, PSNR, SSIM) on each split of CelebA-Test for the comparisons of previous methods and our methods. Detailed quantitative comparisons on total 9 splits are presented here in \cref{Tab:A5}.

\begin{table}[h]
  \centering
    \resizebox{\textwidth}{69mm}{
      \begin{tabular}{c|c|c c c c|c c c c|c c c c}
        \toprule
        ~ & ~ & \multicolumn{4}{c|}{$\times16$} & \multicolumn{4}{c|}{$\times8$} & \multicolumn{4}{c}{$\times4$} \\
         Methods & splits & LPIPS$\downarrow$ & FID$\downarrow$ & PSNR$\uparrow$ & SSIM$\uparrow$ & LPIPS$\downarrow$ & FID$\downarrow$ & PSNR$\uparrow$ & SSIM$\uparrow$ & LPIPS$\downarrow$ & FID$\downarrow$ & PSNR$\uparrow$ & SSIM$\uparrow$ \\
        \midrule
        \multirow{4}*{DFDNet*}    &  mild  & 0.4328 &  68.50 & 23.66 & 0.5959 & 0.3451 &  58.96 & 25.55 & 0.6371 & 0.2733 &  56.22 & 27.47 & 0.7054 \\
        ~                         & medium & 0.5713 & 103.35 & 20.71 & 0.4835 & 0.5293 & 118.01 & 21.47 & 0.4580 & 0.4605 &  93.92 & 23.62 & 0.5317 \\
        ~                         & severe & 0.6491 & 156.39 & 18.04 & 0.3994 & 0.6356 & 183.42 & 18.22 & 0.3322 & 0.5878 & 144.16 & 20.33 & 0.3699 \\
        ~                         & \textbf{avg.}   & \textbf{0.5511} & \textbf{109.41} & \textbf{20.80} & \textbf{0.4929} & \textbf{0.5033} & \textbf{120.13} & \textbf{21.75} & \textbf{0.4758} & \textbf{0.4405} &  \textbf{98.10} & \textbf{23.81} & \textbf{0.5357} \\
        \midrule
        \multirow{4}*{GPEN}       &  mild  & 0.3443 &  53.56 & 23.44 & 0.6428 & 0.2970 & 50.13 & 25.61 & 0.6757 & 0.2326 & 49.44 & 27.77 & 0.7366 \\
        ~                         & medium & 0.4387 &  72.52 & 21.68 & 0.5938 & 0.3666 & 53.82 & 24.17 & 0.6560 & 0.2843 & 50.03 & 26.59 & 0.7225 \\
        ~                         & severe & 0.5108 & 118.64 & 20.20 & 0.5381 & 0.4599 & 88.05 & 22.28 & 0.5878 & 0.3632 & 61.20 & 24.78 & 0.6579 \\
        ~                         & \textbf{avg.}   & \textbf{0.4313} &  \textbf{81.57} & \textbf{21.77} & \textbf{0.5916} & \textbf{0.3745} & \textbf{64.00} & \textbf{24.02} & \textbf{0.6398} & \textbf{0.2934} & \textbf{53.56} & \textbf{26.38} & \textbf{0.7057} \\
        \midrule
        \multirow{4}*{GFP-GAN}    &  mild  & 0.4123 &  53.36 & 21.33 & 0.5810 & 0.2558 & 55.79 & 25.46 & 0.7089 & 0.2065 & 55.76 & 27.45 & 0.7596 \\
        ~                         & medium & 0.5708 & 141.49 & 17.78 & 0.4286 & 0.3205 & 52.55 & 23.37 & 0.6774 & 0.2704 & 54.76 & 25.28 & 0.7264 \\
        ~                         & severe & 0.6458 & 222.53 & 15.95 & 0.3638 & 0.3935 & 62.29 & 21.25 & 0.6221 & 0.3390 & 65.83 & 22.10 & 0.6871 \\
        ~                         & \textbf{avg.}   & \textbf{0.5430} & \textbf{139.13} & \textbf{18.35} & \textbf{0.4578} & \textbf{0.3233} & \textbf{56.88} & \textbf{23.36} & \textbf{0.6695} & \textbf{0.2720} & \textbf{58.78} & \textbf{ 24.94} & \textbf{0.7244} \\
        \midrule
        \multirow{4}*{CodeFormer} &  mild  & 0.4058 &  71.39 & 22.67 & 0.5733 & 0.2486 & 62.33 & 26.04 & 0.6994 & 0.2075 & 64.25 & 27.66 & 0.7491 \\
        ~                         & medium & 0.5392 & 115.32 & 19.33 & 0.4340 & 0.3348 & 60.70 & 24.32 & 0.6398 & 0.2597 & 58.51 & 26.32 & 0.7114 \\
        ~                         & severe & 0.6077 & 164.79 & 17.10 & 0.3586 & 0.4561 & 90.63 & 22.08 & 0.5184 & 0.3090 & 61.47 & 25.02 & 0.6591 \\
        ~                         & \textbf{avg.}   & \textbf{0.5176} & \textbf{117.17} & \textbf{19.70} & \textbf{0.4553} & \textbf{0.3465} & \textbf{71.22} & \textbf{24.15} & \textbf{0.6192} & \textbf{0.2587} & \textbf{61.41} & \textbf{26.33} & \textbf{0.7065} \\
        
        \midrule
        \multirow{4}*{SPARNetHD}  &  mild  & 0.3432 &  64.18 & 23.84 & 0.6567 & 0.2695 & 53.90 & 26.09 & 0.7064 & 0.2151 & 52.45 & 27.68 & 0.7511 \\
        ~                         & medium & 0.4392 &  72.30 & 22.16 & 0.6110 & 0.3370 & 55.62 & 24.67 & 0.6799 & 0.2697 & 51.44 & 26.57 & 0.7270 \\
        ~                         & severe & 0.5042 &  94.59 & 20.83 & 0.5665 & 0.4019 & 69.46 & 23.36 & 0.6365 & 0.3066 & 55.72 & 25.51 & 0.6984 \\
        ~                         & \textbf{avg.}   & \textbf{0.4289} &  \textbf{77.02} & \textbf{22.28} & \textbf{0.6114} & \textbf{0.3361} & \textbf{59.66} & \textbf{24.71} & \textbf{0.6743} & \textbf{0.2638} & \textbf{53.20} & \textbf{26.59} & \textbf{0.7255} \\
        \midrule
        \multirow{4}*{VQFR}       &  mild  & 0.5027 &  81.06 & 20.57 & 0.4735 & 0.2793 & 51.58 & 23.99 & 0.6542 & 0.2466 & 53.84 & 24.48 & 0.6773 \\
        ~                         & medium & 0.6631 & 153.24 & 17.25 & 0.3037 & 0.3978 & 50.75 & 22.19 & 0.5647 & 0.3034 & 50.27 & 23.70 & 0.6444 \\
        ~                         & severe & 0.7277 & 223.39 & 15.37 & 0.2370 & 0.5872 & 97.30 & 19.32 & 0.3846 & 0.3782 & 53.06 & 22.38 & 0.5789 \\
        ~                         & \textbf{avg. }  & \textbf{0.6312} & \textbf{152.56} & \textbf{17.73} & \textbf{0.3381} & \textbf{0.4214} & \textbf{66.54} & \textbf{21.83} & \textbf{0.5345} & \textbf{0.3094} & \textbf{52.39} & \textbf{23.52} & \textbf{0.6335} \\
        \midrule
        \multirow{4}*{\textbf{DR2 + SPAR}} &  mild  & 0.3563 & 58.70 & 24.27 & 0.6907 & 0.2813 & 54.52 & 26.41 & 0.7205 & 0.2237 & 51.60 & 27.89 & 0.7541 \\
        ~                                  & medium & 0.3878 & 54.89 & 22.10 & 0.6603 & 0.3367 & 57.38 & 24.64 & 0.6986 & 0.2695 & 51.78 & 26.14 & 0.7226 \\
        ~                                  & severe & 0.4282 & 46.07 & 20.49 & 0.6250 & 0.3474 & 56.97 & 23.30 & 0.6708 & 0.2972 & 50.93 & 24.82 & 0.7021 \\
        ~                                  &  \textbf{avg.}  & \textbf{0.3908} & \textbf{53.22} & \textbf{22.29} & \textbf{0.6587} & \textbf{0.3218} & \textbf{56.29} & \textbf{24.78} & \textbf{0.6966} & \textbf{0.2635} & \textbf{51.44} & \textbf{26.28} & \textbf{0.7263} \\
        \midrule
        \multirow{4}*{\textbf{DR2 + VQFR}} &  mild  & 0.3374 & 53.05 & 22.94 & 0.6550 & 0.2801 & 54.55 & 24.37 & 0.6897 & 0.2566 & 51.96 & 24.55 & 0.6903 \\
        ~                                  & medium & 0.3966 & 45.41 & 21.13 & 0.6212 & 0.3254 & 52.96 & 23.40 & 0.6899 & 0.2945 & 50.00 & 24.18 & 0.6922 \\
        ~                                  & severe & 0.4340 & 43.40 & 19.79 & 0.5905 & 0.3445 & 53.95 & 22.44 & 0.6611 & 0.3196 & 52.27 & 23.40 & 0.6706 \\
        ~                                  &  \textbf{avg.}  & \textbf{0.3893} & \textbf{47.29} & \textbf{21.29} & \textbf{0.6222} & \textbf{0.3167} & \textbf{53.82} & \textbf{23.40} & \textbf{0.6802} & \textbf{0.2902} & \textbf{51.41} & \textbf{24.04} & \textbf{0.6844} \\
        \bottomrule
      \end{tabular}
      }
  \caption{Detailed quantitative comparisons on \textbf{CelebA-Test}. '*' denotes using ground-truth landmarks as input. Our methods show more advantages at more severely degraded splits.}
  \label{Tab:A5}
\end{table}

\subsection{More Qualitative Comparisons}
For more comprehensive comparisons with previous methods on different levels of degraded dataset, we provide qualitative results on each split of CelebA-Test dataset under each upsampling factor in \cref{Fig:A3,Fig:A4,Fig:A5}.

\begin{figure}[h]
\centering
\includegraphics[width=1.0\linewidth, trim=0 8 0 10 ]{cvpr2023-author_kit-v1_1-1/latex/figs/figsup3_more_comparisons_of_CA_x16.pdf}
\caption{\textbf{Qualitative results on CelebA-Test ($\times16$).} Previous methods produce more artifacts when inputs are heavily degraded.}
\label{Fig:A3}
\end{figure}

\begin{figure}[h]
\centering
\includegraphics[width=1.0\linewidth, trim=0 8 0 10 ]{cvpr2023-author_kit-v1_1-1/latex/figs/figsup4_more_comparisons_of_CA_x8.pdf}
\caption{\textbf{Qualitative results on CelebA-Test ($\times8$).} Previous methods produce more artifacts when inputs are heavily degraded.}
\label{Fig:A4}
\end{figure}

\begin{figure}[h]
\centering
\includegraphics[width=1.0\linewidth, trim=0 8 0 10 ]{cvpr2023-author_kit-v1_1-1/latex/figs/figsup5_more_comparisons_of_CA_x4.pdf}
\caption{\textbf{Qualitative results on CelebA-Test ($\times4$).} Our methods produce comparable results with previous arts on mildly degraded data.}
\label{Fig:A5}
\end{figure}

%%%%%%%%% REFERENCES
% {\small
% \bibliographystyle{ieee_fullname}
% \bibliography{egbib}
% }